\documentclass[11pt]{article}

\usepackage[final]{acl}

\usepackage{times}
\usepackage{latexsym}
\usepackage{amsmath,amssymb}

\usepackage[T1]{fontenc}

\usepackage[utf8]{inputenc}

\usepackage{microtype}

\usepackage{inconsolata}

\usepackage{subcaption}
\usepackage{multirow}
\usepackage{booktabs}

\usepackage{graphicx}
\usepackage{colortbl}
\usepackage{latexsym}
\usepackage{array}  
\usepackage{inconsolata}
\usepackage{mathtools}
\usepackage{extarrows}
\usepackage{arydshln}
\usepackage{array}
\usepackage{booktabs}
\usepackage{amsfonts}
\usepackage{mathtools,amssymb}
\usepackage{bbm, dsfont}
\usepackage{bm}
\usepackage{bbm}
\usepackage[ruled,vlined,linesnumbered]{algorithm2e}
\usepackage{enumitem}
\usepackage{tcolorbox}

\definecolor{ourscolor}{RGB}{220,240,255}
\definecolor{tablecolor}{RGB}{255,0,0}
\usepackage{booktabs} 
\usepackage{array}    
\usepackage{graphicx} 
\usepackage{amsmath}  
\usepackage{textcomp} 
\usepackage{fontawesome}
\usepackage{algpseudocode}

\newcommand{\ie}{\textit{i.e.}}
\newcommand{\eg}{\textit{e.g.}}
\usepackage[table,xcdraw]{xcolor}
\usepackage{xcolor}
\usepackage{url}

\newcommand{\fmtnum}[1]{%
  \ifnum\fpeval{#1 < 0} = 1
    \textcolor{red}{$#1$}%
  \else
      \textcolor{green}{$#1$}%
  \fi
}

\newcommand{\infobox}[1]{
    \vspace{-0.18cm}
    \begin{tcolorbox}[
         colback=white!90!gray,     
        colframe=teal!60!black,   
        arc=5pt,                   
        boxsep=5pt,                 
        left=5pt,                  
        right=10pt,                 
        top=2pt,                   
        bottom=3pt,                
        boxrule=0.8pt,              
    ]
    \vspace{-0.1cm}
         \textit{#1}
    \vspace{-0.2cm}
    \end{tcolorbox}
    \vspace{-0.15cm}
}

\title{Visually-Guided Policy Optimization for Multimodal Reasoning}

\author{
 \textbf{Zengbin Wang\textsuperscript{1}}\footnotemark[1], 
 \textbf{Feng Xiong\textsuperscript{1}}\footnotemark[1], 
 \textbf{Liang Lin\textsuperscript{1}}, 
 \textbf{Xuecai Hu\textsuperscript{1}}\footnotemark[2]
 \\
 \textbf{Yong Wang\textsuperscript{1}}\footnotemark[2], 
 \textbf{Yanlin Wang\textsuperscript{2}}, 
 \textbf{Man Zhang\textsuperscript{3}}, 
 \textbf{Xiangxiang Chu\textsuperscript{1}}
\\ 
 \textsuperscript{1}AMAP, Alibaba Group 
 \quad 
 \textsuperscript{2}SYSU
 \quad 
 \textsuperscript{3}BUPT
}

\begin{document}
\maketitle
{
\renewcommand{\thefootnote}{\fnsymbol{footnote}}
\footnotetext[1]{\ Equal contribution.}
\footnotetext[2]{\ Project leads and corresponding authors.}
}
\begin{abstract}
Reinforcement learning with verifiable rewards (RLVR) has significantly advanced the reasoning ability of vision-language models (VLMs). 
However, the inherent text-dominated nature of VLMs often leads to insufficient visual faithfulness, characterized by sparse attention activation to visual tokens. More importantly, our empirical analysis reveals that temporal visual forgetting along reasoning steps exacerbates this deficiency.
To bridge this gap, we propose \underline{V}isually-\underline{G}uided \underline{P}olicy \underline{O}ptimization (\textbf{VGPO}), a novel framework to reinforce visual focus during policy optimization. 
Specifically, VGPO initially introduces a \textit{Visual Attention Compensation} mechanism that leverages visual similarity to localize and amplify visual cues, while progressively elevating visual expectations in later steps to counteract visual forgetting.
Building on this mechanism, we implement a dual-grained advantage re-weighting strategy: the intra-trajectory level highlights tokens exhibiting relatively high visual activation, while the inter-trajectory level prioritizes trajectories demonstrating superior visual accumulation.
Extensive experiments demonstrate that VGPO achieves better visual activation and superior performance in mathematical multimodal reasoning and visual-dependent tasks.
The code has been released at \url{https://github.com/wzb-bupt/VGPO}.
\end{abstract}


\section{Introduction}
    Vision-Language Models (VLMs)~\cite{hurst2024gpt,bai2025qwen25vl,comanici2025gemini}, renowned for their multimodal reasoning over both image and text, have achieved remarkable progress across multimodal tasks~\cite{lu2024mathvista,zhang2024MathVerse,yue-etal-2025-mmmu,zhougeomath,qiao-etal-2025-math}.
    Recently, the Reinforcement Learning with Verifiable Rewards (RLVR)~\cite{shao2024deepseekmath,yu2025dapo,zheng2025gspo,chu2025gpg,dai2026harder,ji2025tree} has further advanced into complex step-by-step reasoning, yielding enhanced logical coherence and superior performance~\cite{huang2025visionr1,meng2025mmeureka,wang2025vlrethinker,huang2025spotlight,yang2025r1onevision,wang2025perception}.

    \begin{figure}
        \centering
        \includegraphics[width=0.95\linewidth]{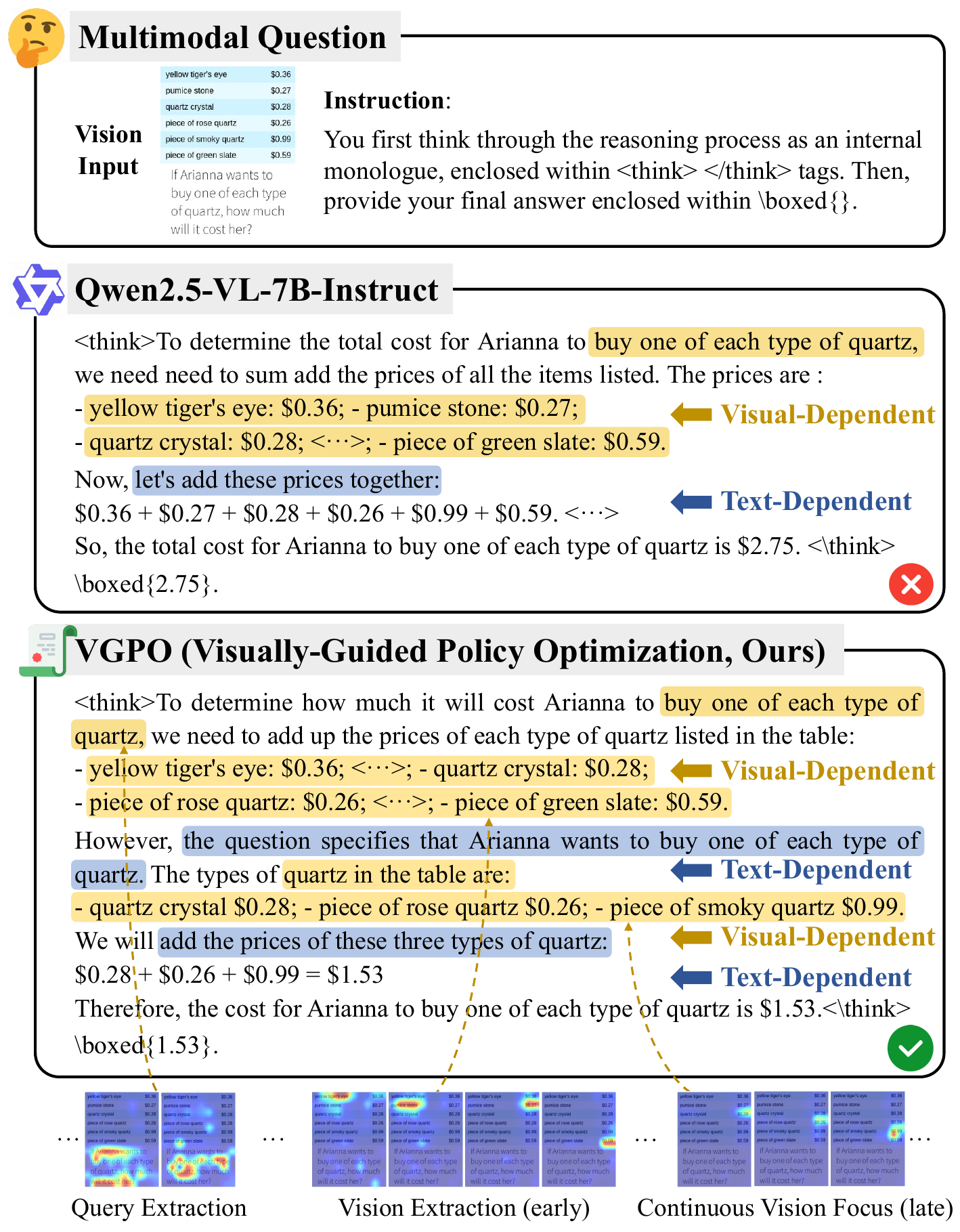}
        \vspace{-0.2cm}
        \caption{A multimodal reasoning example with visual input. Our VGPO not only attends to the query and visual elements in the early reasoning steps, like the base model~\cite{bai2025qwen25vl}, but also exhibits better visual activation and sustains visual focus in the later steps to ensure consistent visual faithfulness.}
        \label{fig01:example}
    \end{figure}

\begin{figure*}[t]
        \centering
        \includegraphics[width=\linewidth]{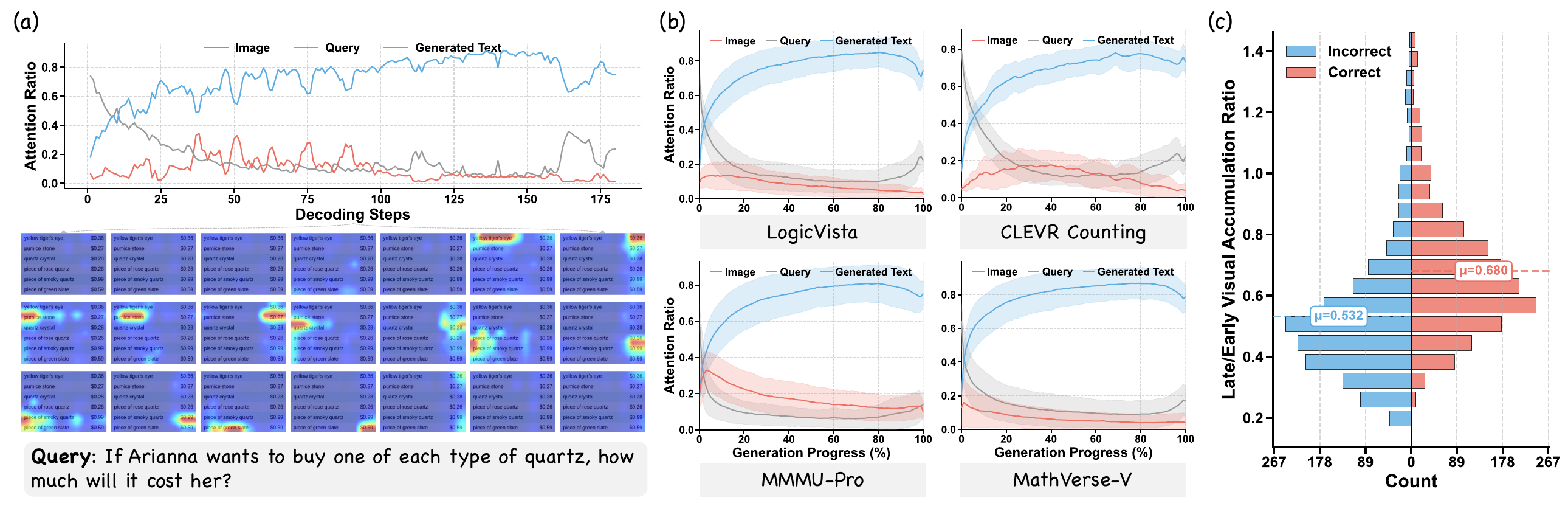}
        \vspace{-0.7cm}
        \caption{Analysis of the inference nature of multimodal reasoning trajectory (based on Qwen2.5-VL-7B~\cite{bai2025qwen25vl}). \textbf{(a)} An example of attention allocation across image, query, and generated text tokens (normalized to 1 at each step). \textbf{(b)} Average attention statistics on four visual-dependent benchmarks~\cite{xiao2024logicvista,li2023super,yue-etal-2025-mmmu,zhang2024MathVerse}. \textbf{(c)} Distribution of late/early visual accumulation ratios for incorrect (left) \textit{vs.} correct (right) samples of these four benchmarks. Incorrect samples often exhibit higher visual forgetting.}
        \label{fig02:analysis}
    \end{figure*}

    Despite exhibiting better logical reasoning, the inherent text-dominated inference nature of VLMs often leads to \textit{insufficient visual faithfulness}~\cite{wang2024picture,fu2024blink,liu2025faithfulness,liu2025seeing}.  
    As discussed in recent literature~\cite{jian2025lookagain,yang2025lookback,favero2024multi} and evidenced by Figures~\ref{fig01:example} and~\ref{fig02:analysis}, models often assign lower attention to input image tokens compared to query tokens and generated text tokens, resulting in sparse activation of input image. 
    More critically, this deficit will be exacerbated by \textit{temporal visual forgetting}: as reasoning chains extend, attention to visual inputs progressively decays. 
    This over-reliance on textual priors rather than visual facts often leads to hallucinations or reasoning errors~\cite{sun2025mitigating,wang2025perception,tian2025VAPO}. Therefore, a natural question arises:
    
    \infobox{Q1: How can we effectively \textit{amplify visual activation} and \textit{mitigate temporal visual forgetting} to ensure visual faithfulness?}
    
    Recent works primarily focused on the former aspect ``amplify visual response'' through external interventions. For example,
    (1) Look-Back~\cite{yang2025lookback} and latent reasoning methods~\cite{li2025lvr,sun2025latentCoT,yang2025mirage} aimed to introduce external learnable special tokens (\eg, <back>, <latent\_strat>, <latent\_end>) that act as explicit triggers to revisit or reconstruct visual input through initial supervised fine-tuning and subsequent autonomous generation.
    (2) VAPO~\cite{tian2025VAPO} explored an additional GPT-5 model to verify the visual faithfulness during intermediate steps and re-activate visual responses through probing visual-related questions.
    (3) PAPO~\cite{wang2025perception} and VPPO~\cite{huang2025spotlight} identified and highlighted visual tokens by contrasting the KL divergence between original versus noisy images through two forward processes.
    While effective, these methods inevitably introduce extra training tokens, auxiliary models, or additional forward passes. In this case, another question arises:

    \infobox{Q2: How can we leverage the model's own internal states to localize visual activation, without any external dependencies?}

    To comprehensively answer these two questions, the key solution lies in (i) precisely localizing visual activations; (ii) amplifying visual activations; and (iii) sustaining visual expectation in the later reasoning steps.
    For the first part, our empirical experiment in Figure~\ref{fig02:analysis} reveals that the inherent hidden states similarity between generated tokens and input image tokens serves as a precise \textit{Visual Focus Score}, enabling spontaneous localization of visually grounded tokens without external supervision.
    For the second and third parts, visual focus score serves as the basis for further designing a \textit{Visual Attention Compensation} mechanism that re-weights visually grounded tokens and progressively improves visual expectations in later reasoning steps to combat temporal visual forgetting.

    Building on the above two insights, we propose \textbf{V}isually-\textbf{G}uided \textbf{P}olicy \textbf{O}ptimization (\textbf{VGPO}). Relying solely on the model's self-generated hidden states to localize visual tokens, VGPO integrates the Visual Attention Compensation mechanism into policy optimization through a \textit{dual-grained advantage re-weighting} strategy. 
    At \textit{intra-trajectory} level, it dynamically re-weights token advantages to amplify specific visual cues, while applying progressive incentives to compensate for temporal visual decay in later steps.
    At \textit{inter-trajectory} level, it globally prioritizes superior visual accumulation, effectively compensating for holistic visual negligence by favoring trajectories that sustain consistent grounding.
    Extensive experiments across mathematical and vision-dependent multimodal reasoning tasks demonstrate that VGPO achieves better visual activation and less visual forgetting.

    In summary, our contributions are four-fold.
    \begin{itemize}[left=0pt]
        \vspace{-0.15cm}
        \item We reveal that the inherent hidden states similarity between generated tokens and image tokens serves as a reliable \textit{Visual Focus Score} to localize visual tokens, thereby facilitating targeted attention modulation without external supervision.
        \vspace{-0.25cm}
        \item We design a \textit{Visual Attention Compensation} mechanism that leverages visual focus score to amplify visual cues, while progressively elevating visual expectations in the latter reasoning step to counteract temporal visual forgetting.
        \vspace{-0.25cm}
        \item We propose VGPO, which integrates the visual attention compensation mechanism into policy optimization using a \textit{Dual-Grained Advantage Re-Weighting} strategy to amplify visual activation and mitigate visual forgetting across intra-trajectory and inter-trajectory levels.
        \vspace{-0.25cm}
        \item Extensive experiments demonstrate that VGPO achieves better visual activation and state-of-the-art performance on mathematical and visual-dependent multimodal reasoning tasks.
    \end{itemize}

\section{Preliminary}
    \subsection{Group Relative Policy Optimization}
    Group Relative Policy Optimization (GRPO)~\cite{shao2024deepseekmath} is a resource-efficient variant of reinforcement learning, which eliminates the value model~\cite{schulman2017ppo} by estimating advantages through group-based computation. 
    Formally, we consider visual input $(I, q, a)$, comprising a visual input $I$, a textual query $q$, and the corresponding ground truth answer $a$, the algorithm begins by sampling a group of $\mathcal{G}$ candidate responses $\{o_{i}\}_{i=1}^{\mathcal{G}}$ from the current policy $\pi_{\theta_{\text{old}}}$.
    For each response $o_i$, a binary reward $r_i \in \{0, 1\}$ is assigned based on the exact match with the ground truth $a$.
    Subsequently, the advantage $\hat{A}_i$ for each response is computed by normalizing the assigned rewards within the sampled group: $\hat{A}_{i} = \frac{r_i - \text{mean}(\{r_k\}_{k=1}^\mathcal{G})}{\text{std}(\{r_k\}_{k=1}^\mathcal{G})}.$
    
    The policy $\pi_{\theta}$ is then updated to maximize the following surrogate objective:
    \begin{equation}
        \label{eq:grpo_objective}
        \small
        \begin{aligned}
            \mathcal{J}_\text{GRPO}(\theta) = & \mathbb{E}_{\{ o_i \}_{i=1}^{\mathcal{G}} \sim \pi_{\theta_\text{old}}( \cdot|I, q)} \Bigg[ \frac{1}{\mathcal{G}}\sum_{i=1}^{\mathcal{G}} \frac{1}{|o_i|}\sum_{t=1}^{|o_i|} \Bigg( \\
            & \hspace{-35pt} \min \Bigg( r_{i,t}(\theta) \hat{A}_{i}, \text{clip} \Big(r_{i,t}(\theta), 1 - \varepsilon, 1 + \varepsilon \Big) \hat{A}_{i} \Bigg) \Bigg],
        \end{aligned}
    \end{equation}
    where $r_{i,t}(\theta) = \frac{\pi_{\theta}(o_{i,t} | q, I, o_{i,<t})}{\pi_{\theta_\text{old}}(o_{i,t} | q, I, o_{i,<t})}$ represents probability ratio. Following DAPO~\citet{yu2025dapo}, we adopt clip-higher setting, asymmetric clipping hyper-parameters, and exclude the KL penalty.

    \subsection{Key Findings in Multimodal Reasoning} \label{sec:key_findings}
    To investigate the underlying flaw in multimodal reasoning, we conduct an empirical analysis of how internal attentions are allocated and evolved throughout VLMs' generation in Figure~\ref{fig02:analysis}. 
    This reveals three key findings that inspire our method.

    \noindent\textbf{(1) Text-dominated Inference and Sparse Visual Activation.}
    First, we analyze the attention allocation among three types of tokens: input image $I$, input query $q$, and current generated text $o_{<t}$. 
    As illustrated in Figure~\ref{fig02:analysis}(a)(b), we observe a significant dominance of textual priors: the model heavily attends to both generated text history (blue line) and input text query (gray line), leaving visual attention (red line) lower than the sum of these text attentions.
    Furthermore, visual activation is often sparsely activated in the early stage, manifesting as a brief ``glance'' the image and followed by long periods of neglect in the later stage.

    \noindent\textbf{(2) Temporal Visual Forgetting.}
    Second, we identify a critical correlation between reasoning length and visual activation, which we term \textit{temporal visual forgetting}. 
    The intensity of visual attention (red line) in Figure~\ref{fig02:analysis}(b) across four benchmarks shows a similar trend: visual activation initially improves and progressively decays as the generation step increases. 
    To further validate the impact of this decay, we statistically compared the attention allocation of correct versus incorrect reasoning samples. 
    Figure~\ref{fig02:analysis}(c) reveals that correct samples often exhibit higher late/early visual accumulation ratios (Average, 0.680 \textit{vs.} 0.532), indicating better temporal visual forgetting mitigation.

    \begin{figure*}[t]
        \centering
        \includegraphics[width=\linewidth]{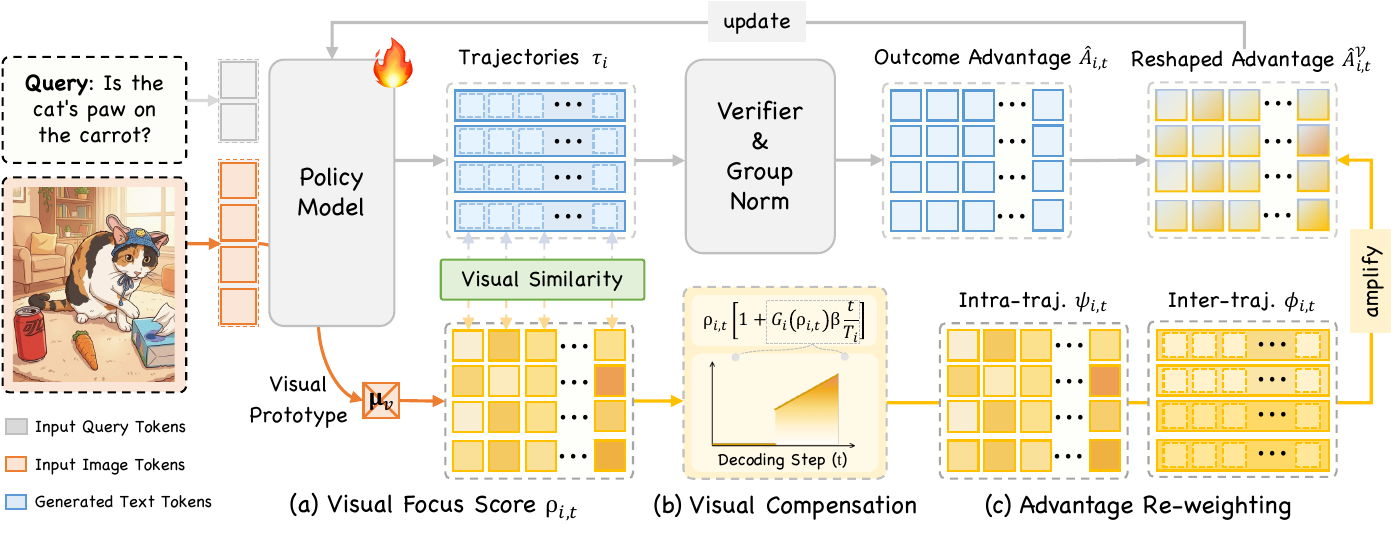}
        \vspace{-0.7cm}
        \caption{
        Overview of \textbf{V}isually-\textbf{G}uided \textbf{P}olicy \textbf{O}ptimization framework. 
        Given query and image, \textbf{(a)} VGPO firstly utilizes the intrinsic hidden state similarity between generated tokens and visual prototype to derive a \textit{Visual Focus Score} for visual token localization.
        \textbf{(b)} Then, \textit{Visual Attention Compensation (VAC)} mechanism leverages this score to re-focus visual tokens, while progressively elevating visual expectations along decoding steps to counteract temporal visual forgetting.
        \textbf{(c)} Finally, \textit{Dual-grained Advantage Re-weighting} strategy integrates VAC mechanism into intra- and inter-trajectory levels to explicitly incentivize sustained visual faithfulness during policy updates.
        }
        \label{fig:method}
    \end{figure*}
    
    \noindent\textbf{(3) Inherent Visual Similarity as a Precise Visual Focus Score.}
    Finally, despite the sparsity of visual activation, we observe a promising phenomenon in the bottom of Figure~\ref{fig02:analysis}(a): when visual tokens are activated (\ie, the peaks in the red line), their attention maps on the original image are highly accurate and semantically grounded based on current base models.
    This indicates the model can locate relevant information but lacks sustained focus in the following reasoning steps.
    Therefore, this visual similarity can serve as a precise, intrinsic \textit{visual focus score} to detect and re-weight critical visual tokens, eliminating the need for auxiliary models.

\section{Methodology}
    In this section, we present Visually-Guided Policy Optimization (VGPO) framework. 
    As in Figure~\ref{fig:method}, VGPO is structured into three key components, including Visual Focus Score (Section~\ref{sec:visual_focus_score}), Visual Attention Compensation mechanism (Section~\ref{sec:visual_attn_compensation}), and Dual-grained Advantage Re-weighting strategy (Section~\ref{sec:dual_reweighting}) to transform intrinsic visual attention into explicit signals to incentivize sustained visual focus throughout the reasoning process.

\subsection{Visual Focus Score} \label{sec:visual_focus_score}
    To quantify the visual engagement of each generated token without external supervision, we introduce an intrinsic metric based on hidden state similarity, as we discussed in Section~\ref{sec:key_findings}.
    Let $\mathcal{H}_v = \{\mathbf{h}^v_{k}\}_{k=1}^{N_v} \subset\mathbb{R}^d$ denote the hidden states of the $N_v$ input image tokens.
    To capture the global visual semantics, we derive a visual prototype $\boldsymbol{\mu}_v$:
    \vspace{-3pt}
    \begin{equation}
        \boldsymbol{\mu}_v = \sum_{k=1}^{N_v} \alpha_k \mathbf{h}^{v}_k, \quad \text{s.t.} \quad \sum \alpha_k = 1,
    \end{equation}
    \vspace{-3pt}
    where $\alpha_k$ represents the weight of the $k$-th image token.
    For the $i$-th trajectory $\tau_i$ at decoding step $t$, we measure the cosine semantic similarity $\mathcal{S}(\cdot, \cdot)$ between the current token hidden state $\mathbf{h}_{i,t}$ and the established visual prototype $\boldsymbol{\mu}_v$:
    \vspace{-3pt}
    \begin{equation}
        \mathcal{S}(\mathbf{h}_{i,t}, \boldsymbol{\mu}_v) = \frac{(\mathbf{h}_{i,t})^\top \boldsymbol{\mu}_v}{\|\mathbf{h}_{i,t}\|_2 \|\boldsymbol{\mu}_v\|_2 + \epsilon},
    \end{equation}
    \vspace{-3pt}
    where $\epsilon$ is a smoothing term. Finally, the visual focus score $\rho_{i,t}$ for the $t$-th token in trajectory $\tau_i$ is obtained via a scaling function to normalize the values to the range $[0, 1]$:
    \begin{equation}
        \rho_{i,t} = \frac{1}{2} \left( \mathcal{S}(\mathbf{h}_{i,t}, \boldsymbol{\mu}_v) + 1 \right).
    \end{equation}

\subsection{Visual Attention Compensation} \label{sec:visual_attn_compensation}
    However, directly employing the raw visual focus score $\rho_{i,t}$ is suboptimal due to the \textit{temporal visual forgetting} phenomenon observed in Section~\ref{sec:key_findings}. 
    As the reasoning chain extends, intrinsic visual attention naturally decays, causing visually grounded tokens in later steps to often exhibit suppressed focus scores and lead to insufficient optimization signals.
    To mitigate this, we introduce a \textit{Visual Attention Compensation} mechanism that counteracts this decay by linearly elevating the expectations in the later reasoning step. 
    Specifically, for $i$-th trajectory $\tau_i$ with token length $T_i$:
    \begin{equation}
        w_{i,t} = \rho_{i,t} \cdot \left[1 + G_i(\rho_{i,t}) \cdot \beta \cdot \frac{t}{T_i}\right],
    \end{equation}
    where the ratio $\frac{t}{T_i}$ indicates a linear strategy to progressively amplify the visual optimization signal as the reasoning step extends.
    $\beta$ is a hyper-parameter controlling {visual compensation intensity}. 
    $G_i(\cdot)$ acts as a visual gate to selectively identify tokens with higher visual similarity (filtering out noisy text tokens) in the later reasoning step.
    Formally, 
    \begin{equation}
        G_i(\rho_{i,t})\hspace{-3pt} =\hspace{-3pt} \begin{cases} 
        1, &\hspace{-8pt}\text{if } t\hspace{-2pt} >\hspace{-2pt} (1\hspace{-2pt}-\hspace{-2pt}\gamma)T_i \ \hspace{-1pt}\text{and}\hspace{-1pt}\  \rho_{i,t}\hspace{-3pt} \ge\hspace{-3pt} \mathcal{Q}_{\kappa}(\mathcal{P}_{\text{tail}, i}) \\ 
        0, &\hspace{-8pt}\text{otherwise} 
        \end{cases}\hspace{-1pt},\hspace{-1pt}
    \end{equation}
    where $\mathcal{P}_{\text{tail}, i}\hspace{-2pt} =\hspace{-2pt} \{\rho_{i,k} \hspace{-2pt}\mid\hspace{-2pt} k\hspace{-2pt} >\hspace{-2pt} (1\hspace{-2pt}-\hspace{-2pt}\gamma)T_i\}$ represents the set of score in the tail of trajectory $\tau_i$, and $\mathcal{Q}_{\kappa}(\cdot)$ returns threshold for the top $\kappa$-percent of this set.
    
\subsection{Dual-grained Advantage Re-weighting} \label{sec:dual_reweighting}
    To effectively integrate visual attention compensation into the policy optimization, we introduce a Dual-grained Advantage Re-weighting mechanism. This approach modulates the standard advantage function by assessing visual relevance at both the \textit{Intra-trajectory} and \textit{Inter-trajectory} levels.
    
    \noindent\textbf{Intra-trajectory Re-weighting.}
    This component aims to capture the distinctions of individual tokens based on their local visual saliency. To highlight visual significance within a specific reasoning path, we derive an intra-trajectory scaling factor $\psi_{i,t}$.
    Specifically, we first normalize the raw visual focus scores $w_{i,t}$ via Min-Max scaling to ensure numerical stability. Subsequently, we zero-center these scores relative to the trajectory's mean, ensuring that only tokens with above-average visual activation are positively incentivized, while effectively suppressing non-visual tokens. Formally,
    \vspace{-2pt}
    \begin{equation}
        \label{eq:token_minmax_norm_1}
        \begin{aligned}
            \widehat{w}_{i,t} &= \frac{{w}_{i,t} - \min_{k} {w}_{i,k}}{\max_{k} {w}_{i,k} - \min_{k} {w}_{i,k} + \epsilon},
            \end{aligned}
            \end{equation}
            \vspace{-2pt}
            \begin{equation}
            \label{eq:token_minmax_norm_2}
            \begin{aligned}
            \psi_{i,t} &= \widehat{w}_{i,t} - \frac{1}{T_i}\sum_{k=1}^{T_i}\widehat{w}_{i,k}.
        \end{aligned}
    \end{equation}
    \vspace{-2pt}
    
    \noindent\textbf{Inter-trajectory Re-weighting.}
    \label{ssec:trajectory_compensation}
    While the intra-trajectory adjustment addresses local granularity, it is equally crucial to evaluate the global visual focus of generated sequences. Correspondingly, we design the inter-trajectory re-weighting strategy to incentivize the model to prioritize entire generations that exhibit superior aggregate visual score.
    For each trajectory $\tau_i$, we compute a cumulative visual score $s_i$, which aggregates the compensated weights over all timesteps. As shown in Eq.~\ref{eq:trajectory_score_decomposed}, this score explicitly accounts for both the base relevance and the time-dependent compensation:
    \begin{equation}
    \small
    \label{eq:trajectory_score_decomposed}
    s_i\hspace{-2pt} = \hspace{-2pt}\sum_{t=1}^{T_i}{w}_{i,t} \hspace{-2pt}= \hspace{-12pt}\underbrace{\sum_{t=1}^{T_i} \rho_{i,t}}_{\text{Visual Relevance}}\hspace{-10pt} + \hspace{-2pt}\underbrace{\sum_{t=1}^{T_i} \rho_{i,t} \hspace{-2pt}\cdot\hspace{-2pt} G_i(\rho_{i,t}) \hspace{-2pt}\cdot\hspace{-2pt} \left(\beta \hspace{-2pt}\cdot\hspace{-2pt} \frac{t}{T_i}\right)}_{\text{Late-stage Compensation}}.
    \end{equation}
    Finally, to derive the final inter-trajectory scaling factor $\phi_i$, we apply group-wise normalization and zero-centering across the rollout group $\mathcal{G}$:
    \begin{equation}
        \label{eq:traj_minmax_norm_1}
        \begin{aligned}
            \widehat{s}_i &= \frac{s_i - \min_{\tau_j \in \mathcal{G}} s_j}{\max_{\tau_j \in \mathcal{G}} s_j - \min_{\tau_j \in \mathcal{G}} s_j + \epsilon},
        \end{aligned}
    \end{equation}
    \vspace{-4pt}
    \begin{equation}
        \label{eq:traj_minmax_norm_2}
        \begin{aligned}
            \phi_i &= \widehat{s}_i - \frac{1}{\mathcal{G}}\sum_{j=1}^\mathcal{G}\widehat{s}_j.
        \end{aligned}
    \end{equation}

\noindent\textbf{Overall Integration.}
\label{ssec:overall_objective}
    By synergizing the intra-trajectory re-weighting with the inter-trajectory re-weighting, we formulate the final visual-focus advantage $\hat{A}_{i,t}$.
    This mechanism essentially modulates the standard outcome-based advantage $\hat{A}_i$ (Eq.~\ref{eq:grpo_objective}) to explicitly incorporate visual focus signals into the policy update.
    The final advantage is computed as a multiplicative integration of the base advantage and the dual-grained visual factors:
    \begin{equation}
    \label{eq:final_advantage}
    \hat{A}_{i,t}^{\mathcal{V}} = \hat{A}_i \cdot (1 + \psi_{i,t}) \cdot (1 + \phi_{i}).
    \end{equation}
    Here, the modulation terms $(1 + \psi_{i,t})$ and $(1 + \phi_{i})$ serve as fine-grained and coarse-grained scaling factors, respectively.
    This composite advantage function reshapes the optimization landscape, ensuring that the policy is driven not merely by the correctness of the final answer, but also by the faithful utilization of visual focus.

\section{Experiments}
    In this section, we present the experimental methodology and the analysis of the results. Further details and results are provided in the Appendix~\ref{sec:appendix_experimental_settings}.

\begin{table*}[t]
    \setlength{\tabcolsep}{1.0mm}
    \renewcommand\arraystretch{1.1}
    \centering
    \resizebox{1.0\linewidth}{!}{
        \begin{tabular}{@{}lcccccccccccc}
        \toprule
        {\multirow{2}{*}{Models}} & \multicolumn{6}{c}{General Mathematical \& Geometric Reasoning} & {\multirow{2}{*}{Avg-$_\mathrm{Math}$}} & \multicolumn{4}{c}{Vision-dependent Multimodal Reasoning} & {\multirow{2}{*}{Avg-$_\mathrm{Vision}$}} \\ 
        \cmidrule{2-7} 
        \cmidrule{9-12}
         & {MathVista} & {MathVerse} & {WeMath} & {MMK12} & {GeoMath} & {Geo3k} &  & {LogicVista} & {Counting} & {MMMU-Pro} & {MathVerse${_V}$} &  \\ 
        \midrule
        \multicolumn{13}{c}{\textit{\textbf{Open-Source General MLLMs (32B-72B)}}} \\ 
        \midrule
        {Qwen2.5-VL-72B} & 74.8 & 65.2$^\dag$ & 69.4$^\dag$ & 65.9$^\dag$ & 53.5$^\dag$ & 54.2$^\dag$ & \cellcolor[HTML]{EFEFEF}{63.8} & 57.9$^\dag$ & 80.5$^\dag$ & 51.1 & 57.6 & \cellcolor[HTML]{EFEFEF}{61.8} \\
        {Qwen3-VL-32B} & 83.8 & 71.5$^\dag$ & 75.2 & 63.1$^\dag$ & 45.7$^\dag$ & 65.4$^\dag$ & \cellcolor[HTML]{EFEFEF}{67.5} & 62.2 & 95.5$^\dag$ & 65.3 & 76.8 & \cellcolor[HTML]{EFEFEF}{75.0} \\ 
        \midrule
        \multicolumn{13}{c}{\textit{\textbf{Open-Source Multimodal Reasoning Models ($\sim$7B)}}} \\ 
        \midrule
        {ThinkLite-VL-7B$^\dag$} & \underline{73.6} & 33.7 & 43.7 & 50.5 & 44.6 & 45.3 & \cellcolor[HTML]{EFEFEF}{48.6} & 44.3 & 88.5 & 32.3 & 30.9 & \cellcolor[HTML]{EFEFEF}{49.0} \\
        {VL-Rethinker-7B$^\dag$} & 65.2 & 68.8 & 67.5 & 66.0 & 50.1 & 44.4 & \cellcolor[HTML]{EFEFEF}{60.3} & 47.0 & 76.5 & 37.6 & 64.8 & \cellcolor[HTML]{EFEFEF}{56.5} \\
        {MM-Eureka-7B$^\dag$} & 66.3 & 66.7 & 66.6 & 61.7 & 49.6 & 41.1 & \cellcolor[HTML]{EFEFEF}{58.7} & 47.9 & 79.0 & \textbf{40.6} & 63.6 & \cellcolor[HTML]{EFEFEF}{57.8} \\
        {NoisyRollout-7B$^\dag$} & 72.9 & 58.8 & 64.4 & 51.2 & 51.6 & \textbf{55.2${^*}$} & \cellcolor[HTML]{EFEFEF}{59.0} & 47.0 & 88.5 & 38.2 & 51.7 & \cellcolor[HTML]{EFEFEF}{56.3} \\
        {$\text{PAPO}_D$-7B$^\dag$} & 72.3 & 69.5 & 69.4 & 80.8 & 52.3 & \underline{48.8} & \cellcolor[HTML]{EFEFEF}{65.5} & 45.9 & \underline{89.0} & {40.2} & 66.6 & \cellcolor[HTML]{EFEFEF}{60.4} \\
        {VPPO-RL-7B$^\dag$} & 70.5 & \underline{71.1} & 70.6 & \textbf{81.8} & \underline{53.2} & 46.9 & \cellcolor[HTML]{EFEFEF}{\underline{65.7}} & 48.8 & 88.5 & {40.2} & \textbf{67.6} & \cellcolor[HTML]{EFEFEF}{\underline{61.3}} \\
        \midrule
        {Qwen2.5-VL-7B} & {68.5} & {40.2} & {47.8} & {49.4} & {51.2} & {42.9} & \cellcolor[HTML]{EFEFEF}{50.0} & {45.2} & {76.5} & {36.4} & {36.6} & \cellcolor[HTML]{EFEFEF}{48.7} \\
        {\ \ + GRPO} & 70.1 & 66.7 & 69.8 & 72.9 & 52.8 & 43.4 & \cellcolor[HTML]{EFEFEF}{62.6\textcolor{tablecolor}{$_{\triangle  25.2\%}$}} & \textbf{49.4} & 85.5 & 38.6 & 61.8 & \cellcolor[HTML]{EFEFEF}{58.8\textcolor{tablecolor}{$_{\triangle  20.7\%}$}} \\
        {\ \ + DAPO} & 68.7 & 69.6 & \underline{70.8} & 77.0 & 51.3 & 45.6 & \cellcolor[HTML]{EFEFEF}{63.8\textcolor{tablecolor}{$_{\triangle  27.6\%}$}} & 47.4 & 85.5 & 39.0 & 66.6 & \cellcolor[HTML]{EFEFEF}{59.6\textcolor{tablecolor}{$_{\triangle  22.4\%}$}} \\
        \rowcolor[HTML]{EFEFEF}
        {\ \ \textbf{+ VGPO (Ours)}} & \textbf{74.1} & \textbf{71.6} & \textbf{72.5} & \underline{81.5} & \textbf{54.3} & 45.8 & \cellcolor[HTML]{EFEFEF}{\textbf{66.6\textcolor{tablecolor}{$_{\triangle  33.2\%}$}}} & \textbf{49.4} & \textbf{95.5} & \underline{40.5} & \textbf{67.6} & \cellcolor[HTML]{EFEFEF}{\textbf{63.3\textcolor{tablecolor}{$_{\triangle  30.0\%}$}}} \\
        \bottomrule
        \end{tabular}
    }
    \vspace{-0.2cm}
    \caption{Performance comparisons across mathematical and vision-dependent multimodal reasoning benchmarks. ``${^*}$'' indicates NoisyRollout is trained on Geo3K. ``$^\dag$'' indicates our reproduction with official checkpoints. Other values are sourced from the official technical report. \textbf{Bold} and \underline{underlined} indicate best and second-best results.}
    \label{fig:main_experiments}
\end{table*}

\subsection{Experimental Settings}

\textbf{Models and Datasets.}
    Consistent with prior studies~\cite{wang2025perception}, we adopt Qwen2.5-VL-series (3B, 7B, and 32B)~\cite{bai2025qwen25vl} as our backbone models.
    We train these models in ViRL39K~\cite{wang2025vlrethinker}, Geo3K~\cite{lu-etal-2021-inter}, and MMK12~\cite{meng2025mmeureka}, while validating in MMK12-val~\cite{meng2025mmeureka}.

\noindent\textbf{Baselines and Evaluations.} 
    We conduct a comprehensive evaluation of our approach against a diverse set of leading VLMs at 7B scale. Our baselines include ThinkLite-VL-7B~\cite{wang2025thinklite-vl}, VL-Rethinker-7B~\cite{wang2025vlrethinker}, MMEureka-7B~\cite{meng2025mmeureka}, NoisyRollout-7B~\cite{liu2025noisyrollout}, $\text{PAPO}_D\text{-}7\text{B}$~\cite{wang2025perception}, and VPPO-RL-7B~\cite{huang2025spotlight}. 
    To assess our VGPO, we employ a suite of benchmarks categorized as follows:
    \noindent(1) \textit{General Mathematical \& Geometric Reasoning.} We utilize MathVista~\cite{lu2024mathvista}, MathVerse~\cite{zhang2024MathVerse}, We-Math~\cite{qiao-etal-2025-math}, MMK12~\cite{meng2025mmeureka}, GeoMath~\cite{zhougeomath}, and Geometry3K~\cite{lu-etal-2021-inter}.
    \noindent(2) \textit{Vision-dependent Multimodal Reasoning.} We adopt LogicVista~\cite{xiao2024logicvista}, SuperClevr Counting~\cite{li2023super}, MMMU-Pro~\cite{yue-etal-2025-mmmu}, and MathVerse-V~\cite{zhang2024MathVerse}. 
    Specifically, we adopt these datasets from PAPO-Eval~\cite{wang2025perception}, which provide filtered subsets enabling robust verification without reliance on LLM-as-a-judge~\cite{chen2024mllm}.
    %

\noindent\textbf{Implementation Details.}
    Following most settings in current methods~\cite{wang2025perception}, we set the training epoch to 2, with learning rate of $1 \times 10^{-6}$, rollout batch size of 512, and maximum response length of 2,048. During the evaluation phase, we set temperature to 0.0 for all experiments. Regarding the specific hyper-parameters of our VGPO, we employ mean-pooling strategy for visual prototype and set compensation intensity $\beta=0.3$, position $\gamma = 0.5$, and $\kappa$-percent $\kappa = 0.2$.

\begin{table*}[t]
    \setlength{\tabcolsep}{1.0mm}
    \renewcommand\arraystretch{1.05}
    \centering
    \resizebox{1.0\linewidth}{!}{
        \begin{tabular}{lcccccccccccc}
        \toprule
        {\multirow{2}{*}{Models}} & \multicolumn{6}{c}{General Mathematical \& Geometric Reasoning} & {\multirow{2}{*}{Avg-$_\mathrm{Math}$}} & \multicolumn{4}{c}{Vision-dependent Multimodal Reasoning} & {\multirow{2}{*}{Avg-$_\mathrm{Vision}$}} \\ 
        \cmidrule{2-7} 
        \cmidrule{9-12}
         & {MathVista} & {MathVerse} & {WeMath} & {MMK12} & {GeoMath} & {Geo3k} &  & {LogicVista} & {Counting} & {MMMU-Pro} & {MathVerse${_V}$} &  \\ 
        \midrule
        \multicolumn{13}{c}{\textit{\textbf{Exploring More Scalable Models (7B $\xrightarrow{}$ 3B, 32B)}}} \\ 
        \midrule
        {Qwen2.5-VL-3B} & 54.0 & 44.1 & 50.2 & 44.1 & 44.3 & 27.6 & \cellcolor[HTML]{EFEFEF}{44.1} & 39.2 & 58.5 & 28.6 & 37.5 & \cellcolor[HTML]{EFEFEF}{41.0} \\
        {\ \ + DAPO} & 63.9 & 57.1 & \textbf{62.9} & 64.1 & 47.6 & \textbf{36.3} & \cellcolor[HTML]{EFEFEF}{55.3\textcolor{tablecolor}{$_{\triangle  25.4\%}$}} & 43.0 & 65.5 & 31.5 & 53.1 & \cellcolor[HTML]{EFEFEF}{48.3\textcolor{tablecolor}{$_{\triangle  17.8\%}$}} \\
        \rowcolor[HTML]{EFEFEF}
        {\ \ \textbf{+ VGPO}} & \textbf{65.0} & \textbf{61.4} & {62.1} & \textbf{71.6} & \textbf{50.2} & {36.1} & \textbf{57.7}\textcolor{tablecolor}{$_{\triangle  30.8\%}$} & \textbf{45.9} & \textbf{78.0} & \textbf{32.5} & \textbf{58.1} & \textbf{53.6}\textcolor{tablecolor}{$_{\triangle  30.7\%}$} \\
        \midrule
        {Qwen2.5-VL-32B} & \textbf{76.8} & 62.4 & 68.8 & 60.0 & 53.7 & 50.8 & \cellcolor[HTML]{EFEFEF}{62.1} & 55.7 & 80.5 & 46.0 & 58.4 & \cellcolor[HTML]{EFEFEF}{60.2} \\
        {\ \ + DAPO} & 69.7 & 73.0 & \textbf{78.7} & 82.0 & 55.2 & 51.6 & \cellcolor[HTML]{EFEFEF}{68.4\textcolor{tablecolor}{$_{\triangle 10.1\%}$}} & 59.1 & 85.5 & \textbf{47.3} & 67.4 & \cellcolor[HTML]{EFEFEF}{64.8\textcolor{tablecolor}{$_{\triangle 7.6\%}$}} \\
        \rowcolor[HTML]{EFEFEF}
        {\ \ \textbf{+ VGPO}} & 75.3 & \textbf{73.8} & \textbf{78.7} & \textbf{87.3} & \textbf{56.5} & \textbf{52.3} & \textbf{70.7}\textcolor{tablecolor}{$_{\triangle 13.8\%}$} & \textbf{59.3} & \textbf{90.0} & \textbf{47.3} & \textbf{70.1} & \textbf{66.7}\textcolor{tablecolor}{$_{\triangle 10.8\%}$} \\
        \midrule
        \multicolumn{13}{c}{\textit{\textbf{Exploring More Training DataSets (39K  $\xrightarrow{}$ 2.1K, 6.4K)}}} \\ 
        \midrule
        {Qwen2.5-VL-7B} & {68.5} & {40.2} & {47.8} & {49.4} & {51.2} & {42.9} & \cellcolor[HTML]{EFEFEF}{50.0} & {45.2} & {76.5} & {36.4} & {36.6} & \cellcolor[HTML]{EFEFEF}{48.7} \\
        {\ \ + DAPO w/ Geo3K $_{\textcolor{tablecolor}{2.1K}}$} & 71.6 & 56.3 & 63.6 & 51.1 & 49.4 & 52.3 & \cellcolor[HTML]{EFEFEF}{57.4\textcolor{tablecolor}{$_{\triangle  14.8\%}$}} & 46.5 & \textbf{85.5} & 36.8 & 50.3 & \cellcolor[HTML]{EFEFEF}{54.8\textcolor{tablecolor}{$_{\triangle 12.5\%}$}} \\
        \rowcolor[HTML]{EFEFEF}
        {\ \ \textbf{+ VGPO w/ Geo3K $_{\textcolor{tablecolor}{2.1K}}$}} & \textbf{72.2} & \textbf{62.5} & \textbf{66.6} & \textbf{53.0} & \textbf{52.6} & \textbf{55.6} & \textbf{60.4\textcolor{tablecolor}{$_{\triangle  20.8\%}$}} & \textbf{47.7} & 80.5 & \textbf{37.0} & \textbf{57.8} & \textbf{55.8\textcolor{tablecolor}{$_{\triangle 14.6\%}$}} \\
        \midrule
        {\ \ + DAPO w/ MMK12 $_{\textcolor{tablecolor}{6.4K}}$} & 72.6 & 66.2 & 67.5 & 64.9 & 51.2 & 42.3 & \cellcolor[HTML]{EFEFEF}{60.8\textcolor{tablecolor}{$_{\triangle  21.6\%}$}} & 47.4 & 89.5 & 37.1 & 61.2 & \cellcolor[HTML]{EFEFEF}{58.8\textcolor{tablecolor}{$_{\triangle 20.7\%}$}} \\
        \rowcolor[HTML]{EFEFEF}
        {\ \ \textbf{+ VGPO w/ MMK12 $_{\textcolor{tablecolor}{6.4K}}$}} & \textbf{74.1} & \textbf{68.5} & \textbf{68.9} & \textbf{66.0} & \textbf{53.2} & \textbf{43.4} & \textbf{62.4}\textcolor{tablecolor}{$_{\triangle 24.8\%}$} & \textbf{48.8} & \textbf{90.0} & \textbf{38.6} & \textbf{63.6} & \textbf{60.3}\textcolor{tablecolor}{$_{\triangle 23.8\%}$} \\
        \bottomrule
        \end{tabular}
    }
    \vspace{-0.2cm}
    \caption{Performance comparisons across different model scales and training datasets.}
    \label{fig:more_model_dataset}
    \vspace{-0.3cm}
\end{table*}

\begin{table}[t]
    \setlength{\tabcolsep}{1mm}
    \centering
    \resizebox{1.0\linewidth}{!}{
        \begin{tabular}{lccc}
        \toprule
        Method & Avg-$_\mathrm{Math}$ & Avg-$_\mathrm{Vision}$ & Overall \\
        \midrule
        {DAPO Baseline} & {63.8} & {59.6} & {62.2} \\
        {\ \ + Intra-trajectory} & 66.1 & 62.5 & 64.6  \\
        {\ \ + Inter-trajectory} & 65.3 & 62.0 & 64.0   \\
        \rowcolor[HTML]{EFEFEF}
        {\ \ + Intra- \& Inter-trajectory} & \textbf{66.6} & \textbf{63.3} & \textbf{65.3} \\
        \bottomrule
        \end{tabular}
    }
    \vspace{-0.2cm}
    \caption{Ablation study on the impact of Intra- and Inter-trajectory re-weighting strategies.}
    \label{table:ablation_module}
\end{table}

\begin{table}[t]
    \setlength{\tabcolsep}{1mm}
    \centering
    \resizebox{1.0\linewidth}{!}{
        \begin{tabular}{lccc}
        \toprule
        Method & Avg-$_\mathrm{Math}$ & Avg-$_\mathrm{Vision}$ & Overall \\
        \midrule
        {Qwen2.5-VL-7B} & 50.0 & 48.7 & 49.5 \\
        {\ \ + DAPO} & 63.8 & 59.6 & 62.2  \\
        {\ \ + DAPO w/ Entropy} &  65.6  & 61.9 &  64.1  \\
        {\ \ + DAPO w/ KL$_{perception}$} & 65.7 & 61.3 & 63.9   \\
        \rowcolor[HTML]{EFEFEF}
        {\ \ + VGPO (Ours)} & \textbf{66.6} & \textbf{63.3} & \textbf{65.3}  \\
        \bottomrule
        \end{tabular}
    }
    \caption{Comparisons with other advantage shaping strategies, including the Entropy-based~\cite{cheng2025entropyreasoning} and KL-based~\cite{huang2025spotlight} methods.}
    \label{table:ablation_adv_shaping}
\end{table}

\begin{figure}[t]
    \centering
    \vspace{-0.2cm}
    \begin{subfigure}{0.48\linewidth}
        \includegraphics[width=\linewidth]{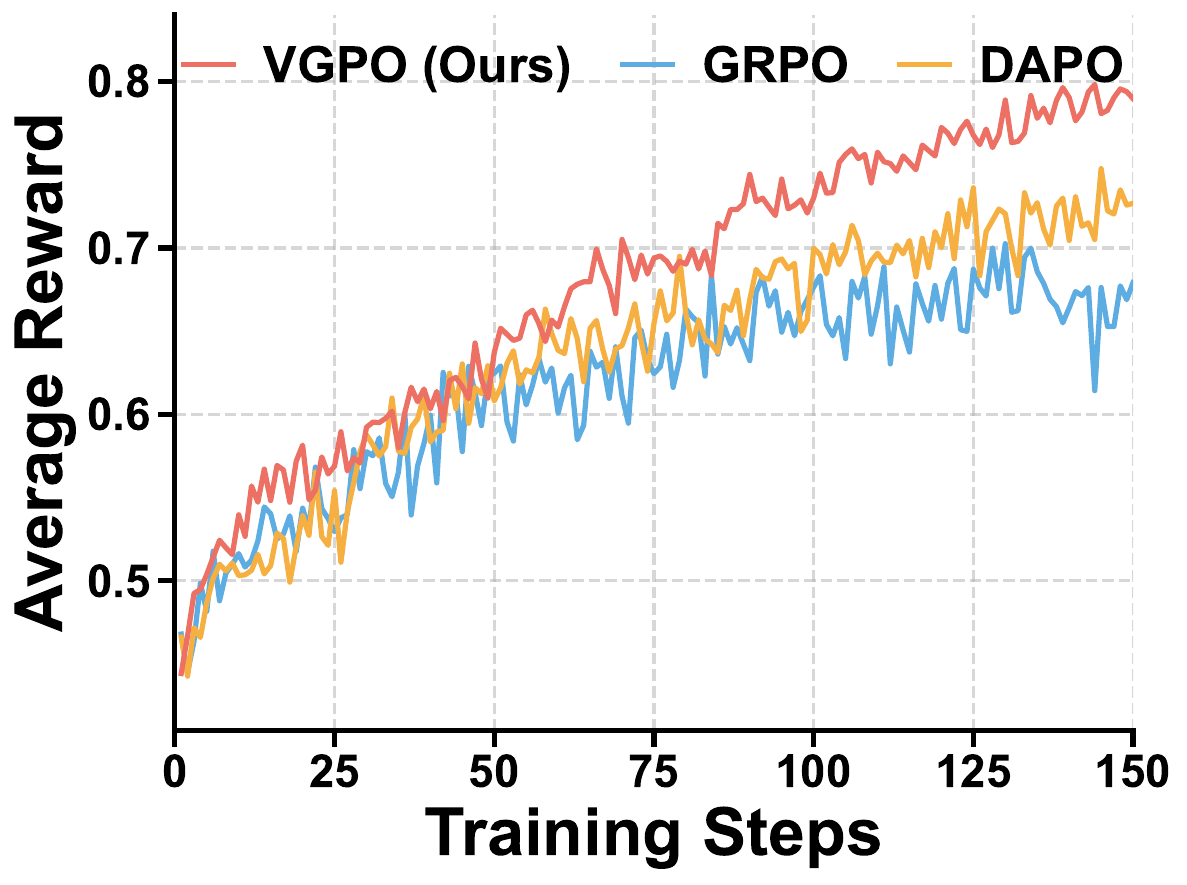}
        \subcaption{Training rewards.}
        \label{fig:training_dynamics_a}
    \end{subfigure}
    \hspace{2pt}
    \begin{subfigure}{0.48\linewidth}
        \includegraphics[width=\linewidth]{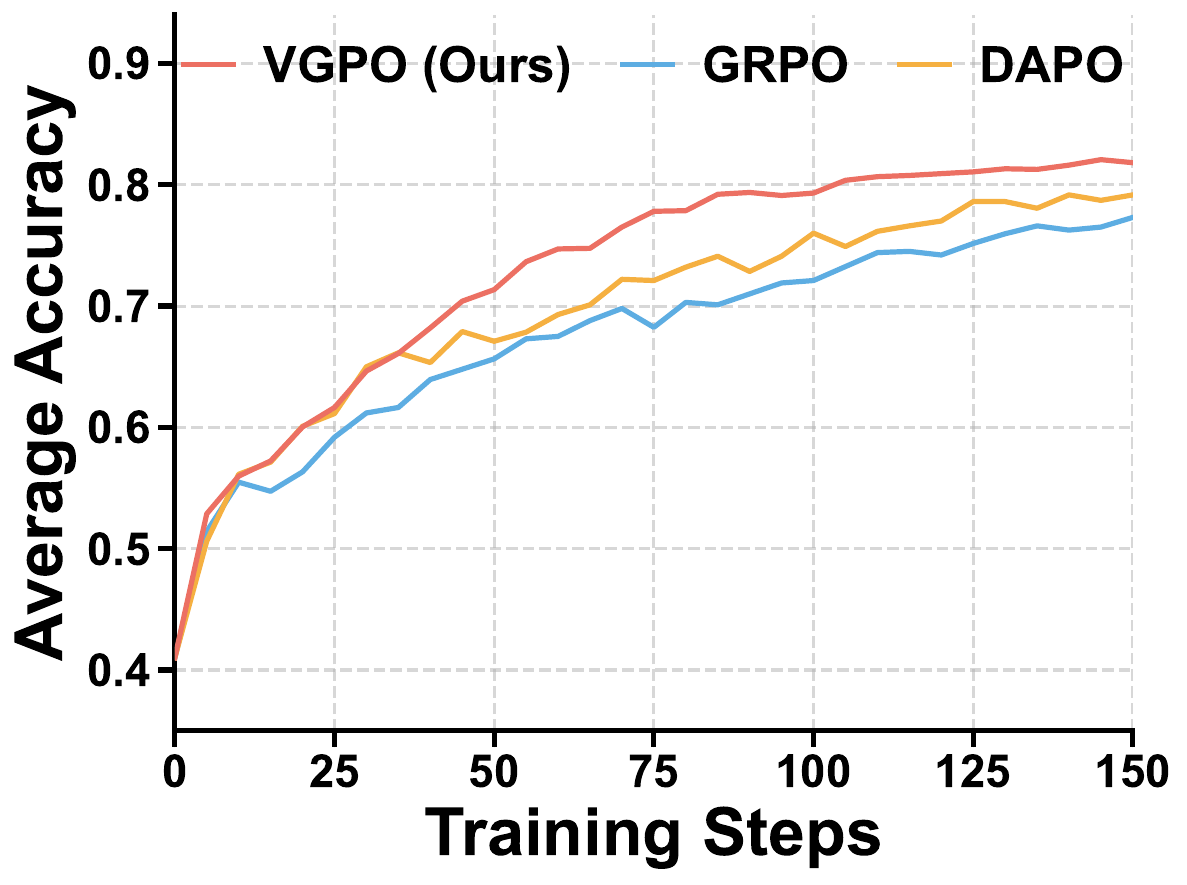}
        \subcaption{Validation accuracy.}
        \label{fig:training_dynamics_b}
    \end{subfigure}
    \vspace{-0.4cm}
    \caption{Training dynamics based on Qwen2.5-VL-7B: \textbf{(a)} training rewards and \textbf{(b)} validation accuracy on MMK12~\cite{meng2025mmeureka} across GRPO~\cite{shao2024deepseekmath}, DAPO~\cite{yu2025dapo}, and our VGPO.}
    \label{fig:training_dynamics}
\end{figure}

\begin{figure}[t]
    \centering
    \begin{subfigure}[t]{0.48\linewidth}
        \includegraphics[width=\linewidth]{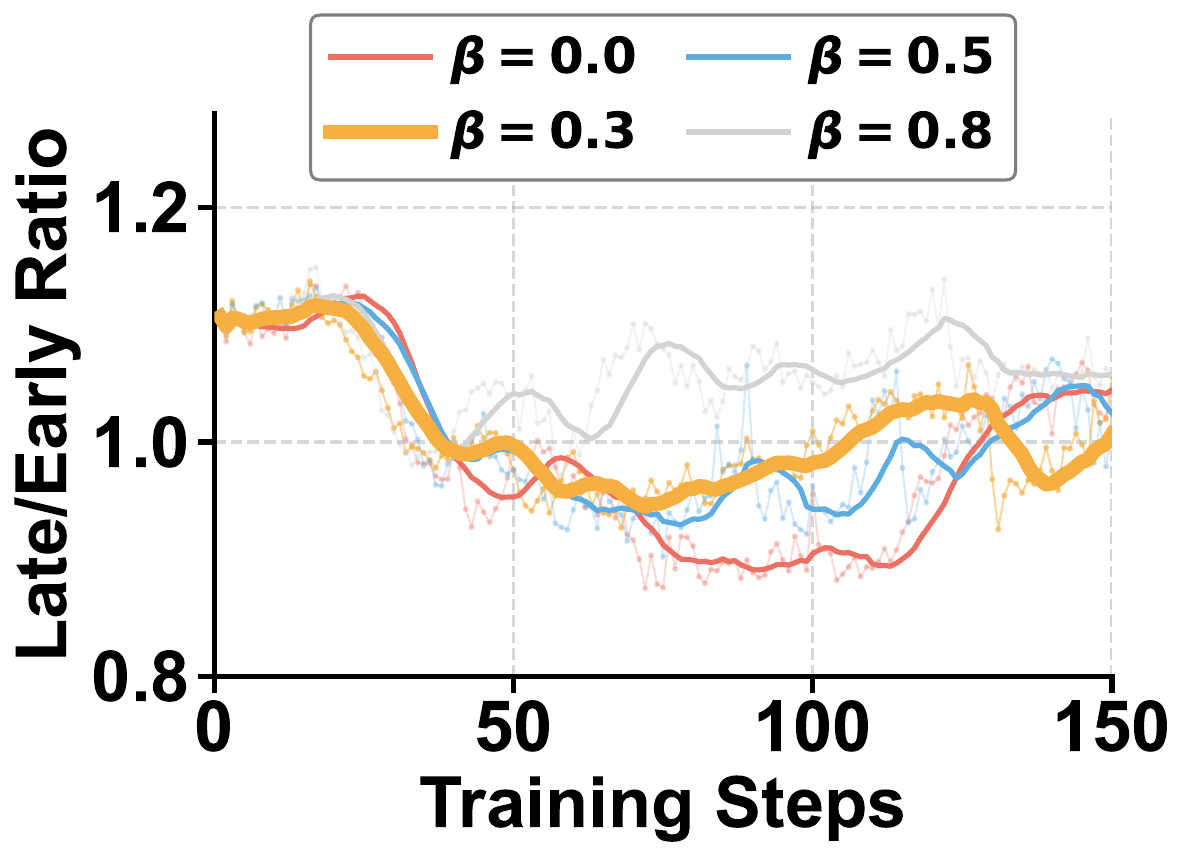}
        \subcaption{Training dynamics of $\beta$.}
    \end{subfigure}
    \begin{subfigure}[t]{0.48\linewidth}
        \includegraphics[width=\linewidth]{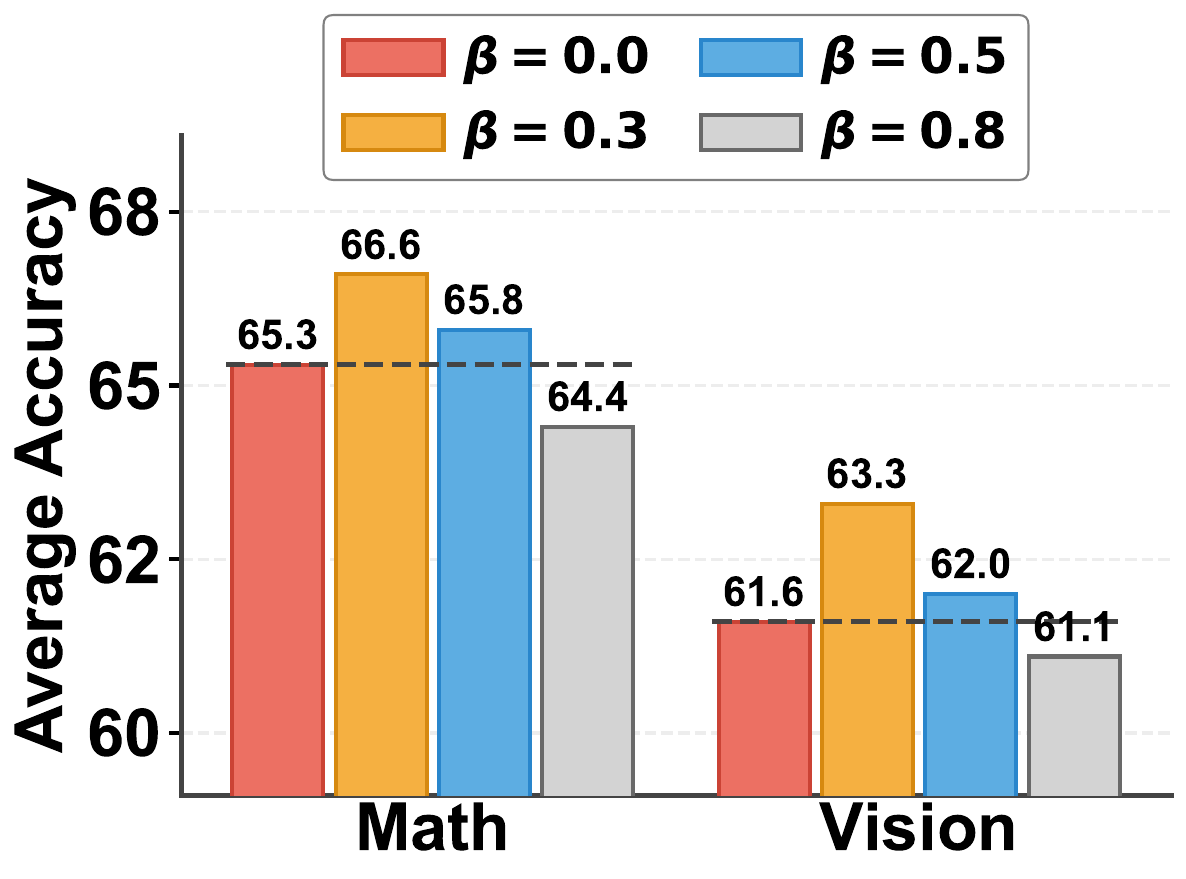}
        \subcaption{Performance w.r.t. $\beta$.}
    \end{subfigure}
    \begin{subfigure}[t]{0.48\linewidth}
        \includegraphics[width=\linewidth]{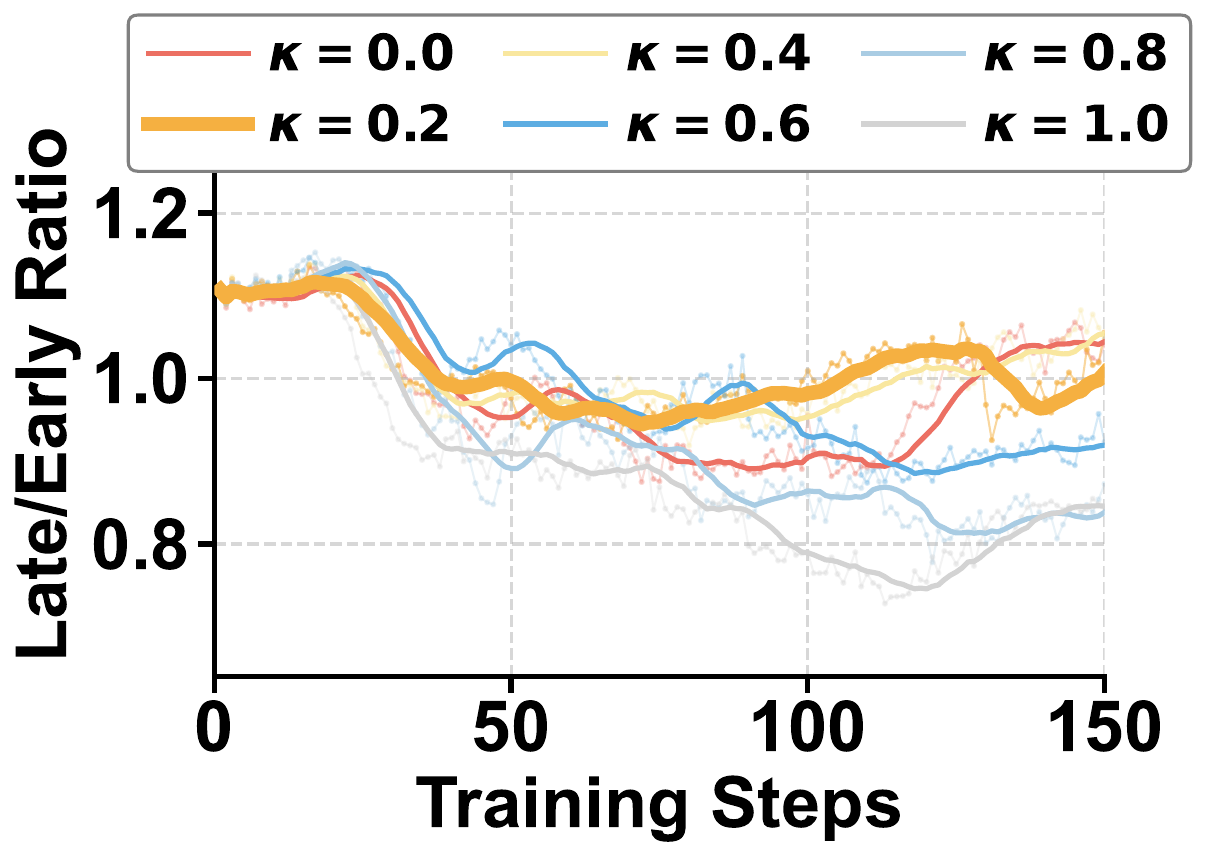}
        \subcaption{Training dynamics of $\kappa$.}
    \end{subfigure}
    \begin{subfigure}[t]{0.48\linewidth}
        \includegraphics[width=\linewidth]{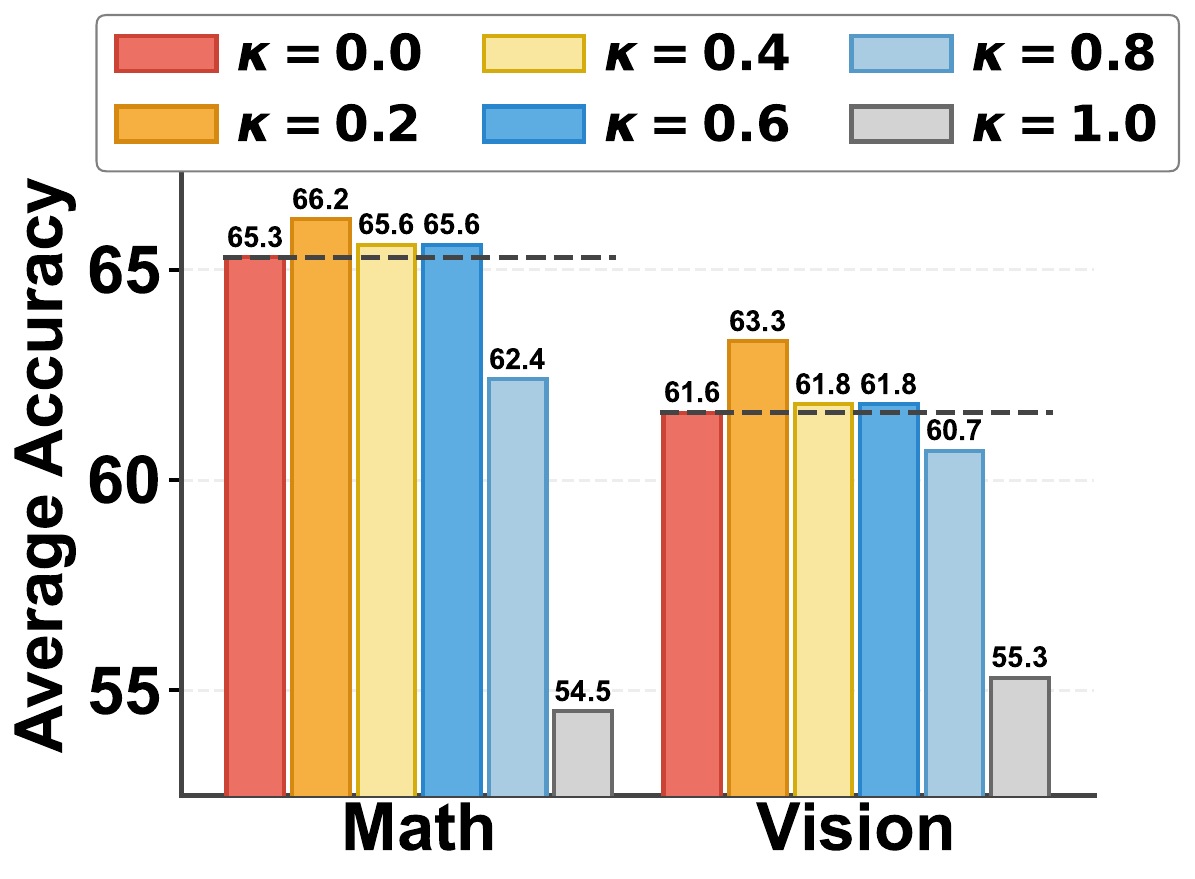}
        \subcaption{Performance w.r.t. $\kappa$.}
    \end{subfigure}
    \begin{subfigure}[t]{0.48\linewidth}
        \includegraphics[width=\linewidth]{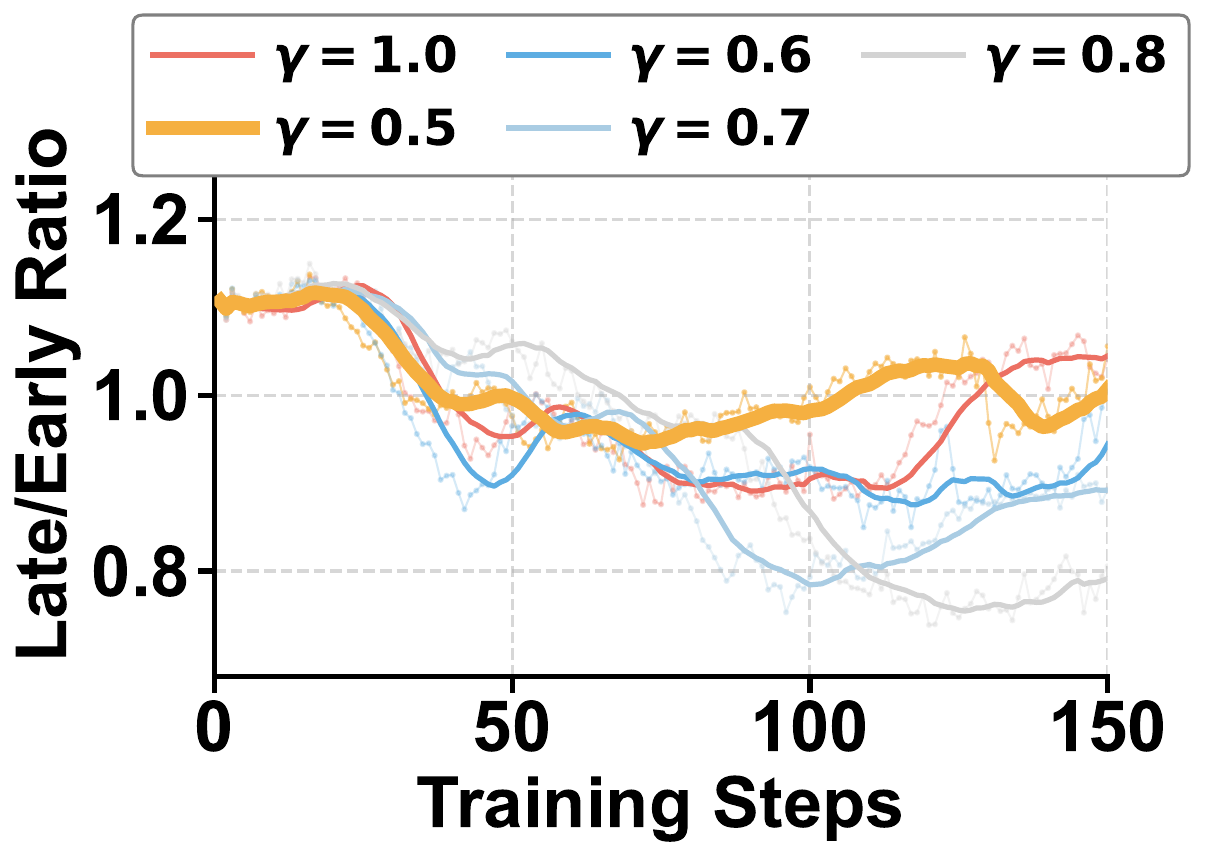}
        \subcaption{Training dynamics of $\gamma$.}
    \end{subfigure}
    \begin{subfigure}[t]{0.48\linewidth}
        \includegraphics[width=\linewidth]{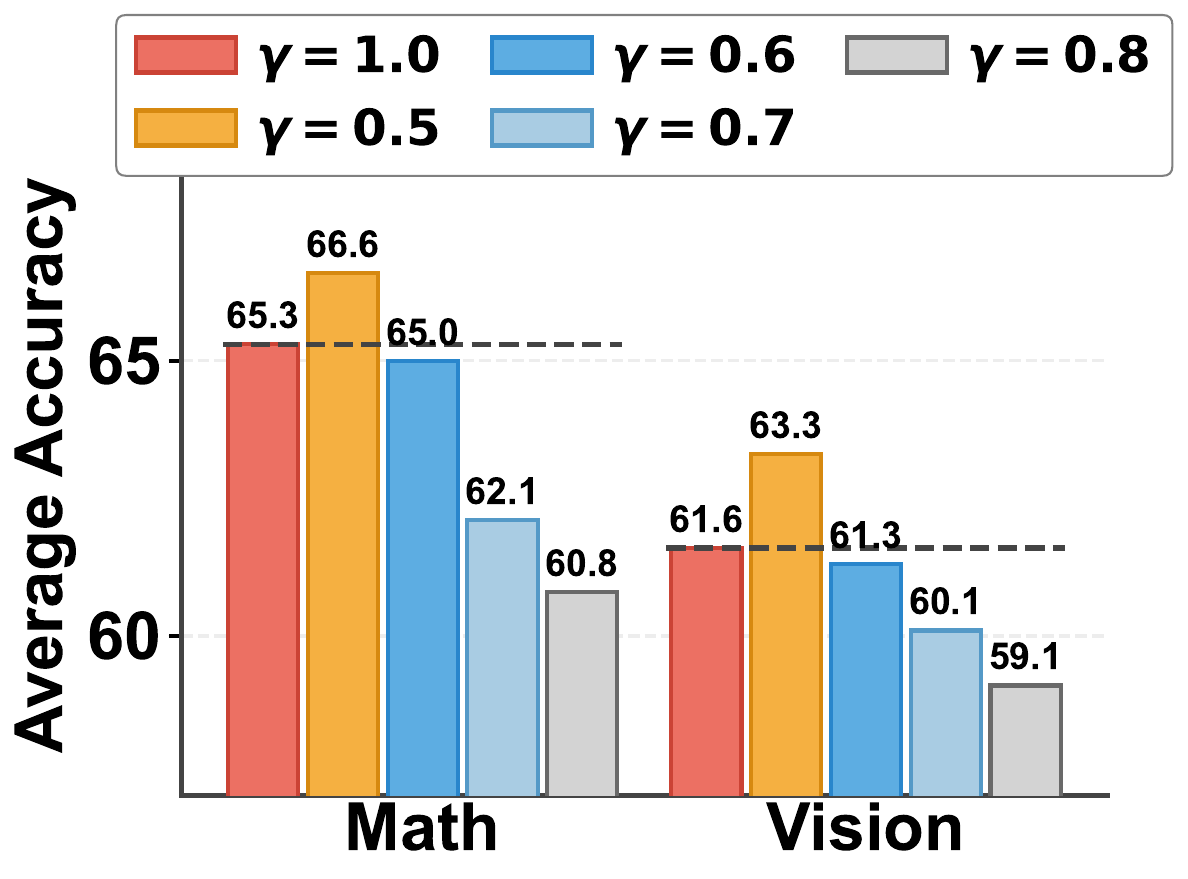}
        \subcaption{Performance w.r.t. $\gamma$.}
    \end{subfigure}
    \caption{Ablation study of training dynamics of the late/early ratio and performance on hyperparameters $\beta$, $\kappa$, and $\gamma$. The Late/Early Ratio is calculated by dividing the visual attention score of the late reasoning stage by that of the early stage.}
    \label{fig:hyper_parameter}
\end{figure}

\subsection{Main Results}

\noindent\textbf{Superior Performance on Multimodal Reasoning.}
    As shown in Table \ref{fig:main_experiments}, our proposed VGPO achieves significant improvements over the base Qwen2.5-VL-7B model, delivering substantial relative gains of 33.2\% on general mathematical reasoning and 30.0\% on vision-dependent multimodal reasoning tasks. 
    Furthermore, when compared to other advanced reasoning models initialized from the same backbone, VGPO consistently secures the top performance, surpassing previous state-of-the-art results with superior average accuracies of 66.6\% in general mathematical benchmarks and 63.3\% in vision-dependent tasks. 
    Notably, despite the significant disparity in model scale, our 7B model exhibits highly competitive performance comparable to the much larger Qwen2.5-VL-72B, highlighting the efficiency of our approach in maximizing the potential of smaller-scale models.

\noindent\textbf{Training Dynamics and Generalization.}
    We further analyze the training stability and generalization capabilities in Figure~\ref{fig:training_dynamics}. 
    VGPO demonstrates a more stable and efficient learning trajectory, achieving consistently higher training rewards compared to baselines like GRPO and DAPO. 
    This superiority can also extend to generalization, where VGPO maintains a clear lead in validation accuracy throughout the training steps. 
    Unlike competitive methods that may suffer from instability or slower convergence, VGPO establishes a robust optimization path, ensuring steady performance gains.

\noindent\textbf{Scalability across Model Scales and Training Data.}
    To validate the scalability, as summarized in Table \ref{fig:more_model_dataset}, experiments ranging from 3B to 32B parameters (Qwen2.5-VL-based) demonstrate that VGPO consistently outperforms baselines, achieving significant relative gains (\eg, 13.8\% on mathematical tasks, 10.8\% on visual-dependent tasks) based on 32B model. 
    Additionally, evaluations on different training datasets with varying sizes reveal superior data efficiency, where our method surpasses the DAPO baseline even in data-constrained regimes. These results suggest that incorporating visual focus into the training process is a generalizable strategy to enhance multimodal reasoning.

\subsection{Quantitative Analysis}

\noindent\textbf{Ablation Study on Re-weighting Strategy.}
    As detailed in Table~\ref{table:ablation_module}, while the baseline model provides a solid foundation, the addition of both intra- and inter-trajectory mechanisms provides extra gains in reliability and specific tasks. 
    Their combination achieves peak average performance across all metrics, confirming that these two strategies are highly complementary and work synergistically to enhance multimodal reasoning capability.

    \begin{table}[t]
    \setlength{\tabcolsep}{3mm}
    \centering
    \resizebox{1.0\linewidth}{!}{
        \begin{tabular}{lccc}
        \toprule
        Method & Avg-$_\mathrm{Math}$ & Avg-$_\mathrm{Vision}$ & Overall \\
        \midrule
        {Baseline (DAPO)} & {63.8} & {59.6} & {62.2} \\
        {\ \ + Step-Function} & 64.7 & 60.7 & 63.1  \\
        {\ \ + Exponential} & 65.1 & 61.0 & 63.5   \\
        \rowcolor[HTML]{EFEFEF}
        {\ \ \textbf{+ Linear (Ours)}} & \textbf{66.6} & \textbf{63.3} & \textbf{65.3}  \\
        \bottomrule
        \end{tabular}
    }
    \vspace{-0.1cm}
    \caption{Ablation study on the impact of different compensation schedules.}
    \label{table:ablation_compensation_schedules}
\end{table}

    \begin{table}[t]
    \setlength{\tabcolsep}{1mm}
    \renewcommand{\arraystretch}{1.1}
    \centering
    \resizebox{1.0\linewidth}{!}{
        \begin{tabular}{lccc}
        \toprule
        Method & Avg-$_\mathrm{Math}$ & Avg-$_\mathrm{Vision}$ & Overall \\
        \midrule
        {Baseline (DAPO)} & {63.8} & {59.6} & {62.2} \\
        {\ \ + Full-trajectory} & 53.0 & 54.2 & 53.5  \\
        \rowcolor[HTML]{EFEFEF}
        {\ \ \textbf{+ Late-trajectory (Ours)}} & \textbf{66.6} & \textbf{63.3} & \textbf{65.3}  \\
        \bottomrule
        \end{tabular}
    }
    \vspace{-0.1cm}
    \caption{Ablation study on the impact of Full-trajectory versus Late-trajectory compensation.}
    \label{table:ablation_full_trajectory}
    \vspace{-0.2cm}
\end{table}

    \begin{figure}[t]
        \centering
        \includegraphics[width=1.0\linewidth]{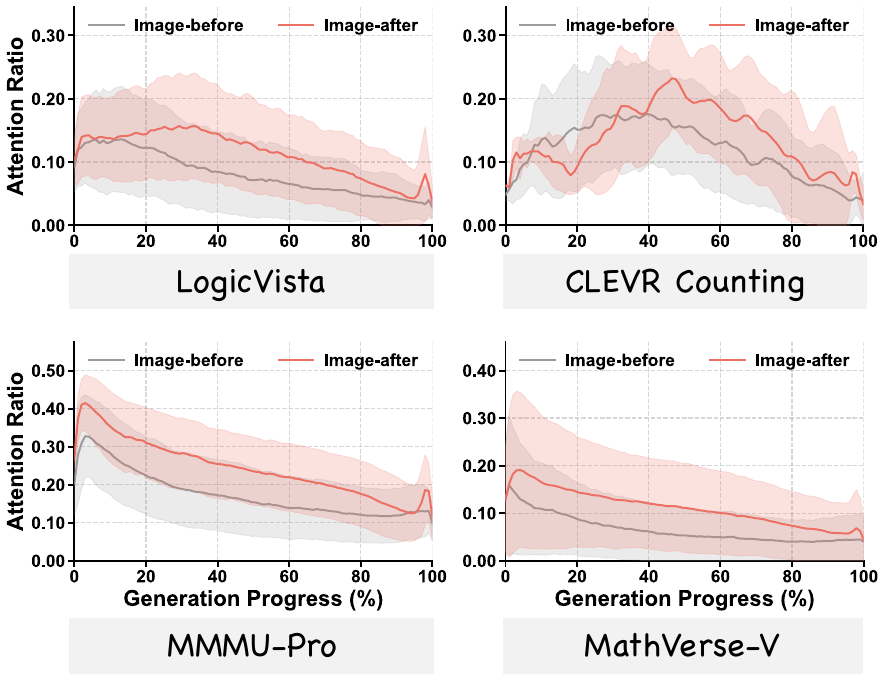}
        \vspace{-0.6cm}
        \caption{Comparison of the vision attention ratio distribution before and after our VGPO across four visual-dependent multimodal reasoning benchmarks.}
        \label{fig06:analysis_visual_attention_ratio}
    \end{figure}

\noindent\textbf{Comparison with Existing Advantage Shaping Methods.}
    We evaluate the efficacy of our VGPO against established advantage shaping strategies, specifically regularized derivatives of DAPO.
    As shown in Table~\ref{table:ablation_adv_shaping}, while entropy and KL regularization effectively regulate advantage estimation relative to vanilla DAPO, VGPO achieves superior empirical performance with a peak average accuracy of 63.3\%.
    This advantage is particularly pronounced in vision-centric tasks like LogicVista and Counting, confirming that strengthening visual focus is a more effective strategy for enhancing multimodal reasoning than generic regularization.

\noindent\textbf{Sensitivity to Hyperparameters.} 
    To evaluate the impact of different parameters on our method and uncover the underlying relation between training dynamics and performance, we conduct a sensitivity analysis on three key hyperparameters: compensation intensity $\beta$, threshold $\kappa$, and tail ratio $\gamma$. 
    As in Figure~\ref{fig:hyper_parameter}, while the model exhibits performance fluctuations across different settings, it achieves optimal accuracy with specific configurations (\ie, $\beta=0.3$, $\kappa=0.2$, and $\gamma=0.5$). 
    Crucially, by correlating these performance peaks with the evolution of the Late/Early Ratio, we identify a decisive factor for training stability: the model yields the best performance when this ratio converges to or stabilizes near 1. 
    This convergence implies that the optimal hyperparameter configuration facilitates a necessary equilibrium between late and early visual focus. 
    By harmonizing the contributions of these two stages, our method ensures robust performance across diverse reasoning tasks.

    \noindent\textbf{Comparison of Different Compensation Schedules.}
    To explore the impact of different visual compensation schedules, we compare our Linear strategy against Step-Function and Exponential schedules, as summarized in Table~\ref{table:ablation_compensation_schedules}. The Linear strategy consistently achieves the best performance. This is because the Exponential schedule tends to over-correct by placing too much emphasis on the final tokens (often formatting or calculation), while the Step-Function schedule introduces training instability by abruptly changing the compensation value. The Linear schedule, however, aligns well with the progressive and continuous decay of visual attention observed in our empirical analysis. Detailed results and analyses are provided in Appendix~\ref{sec:appendix_compensation_schedules}.

    \noindent\textbf{Comparison with Full-trajectory Compensation.}
    We justify our choice of the late-trajectory compensation strategy by comparing it with a full-trajectory approach, as shown in Table~\ref{table:ablation_full_trajectory}. The full-trajectory compensation significantly drops performance across most benchmarks. This is because current VLMs naturally exhibit high visual attention in the early stages, and enforcing additional visual compensation early can distract the model from parsing the textual query or overemphasize early visual attention. Our late-trajectory design specifically targets the later stages where visual decay occurs, achieving superior results. Detailed comparisons are provided in Appendix~\ref{sec:appendix_full_trajectory}.

\noindent\textbf{Analysis of Visual Attention Allocation after our VGPO.}
    To investigate whether our VGPO achieves better visual activation and temporal visual forgetting mitigation, we compare the visual attention ratio of input image before and after our VGPO in Figure~\ref{fig06:analysis_visual_attention_ratio}.
    Our VGPO exhibits higher visual attention allocation throughout the entire generation process and sustains better temporal visual forgetting mitigation compared with the baseline. This prolonged visual grounding is critical for accurate long-chain reasoning, ensuring the model does not lose track of visual evidence in later stages.

\section{Related Work}
    \paragraph{Multimodal Reasoning Challenges.}
    Following the milestone of Large Language Models (LLMs) in complex step-by-step reasoning~\cite{wei2022chain,xia2025beyond,yang2025qwen3}, the research focus has naturally shifted toward extending these abilities to Vision-Language Models (VLMs) for broader real-world applications through integrating vision encoders~\cite{vaswani2017attention} and large language models~\cite{zhang2024visionsurvey,li2025perceptionsurvey,zhu2025internvl3,bai2025qwen3vltechnicalreport}.
    However, despite significant architectural advancements, current VLMs largely inherit the text-dominated inductive biases of their LLM backbones, frequently manifesting as insufficient visual faithfulness and severe visual hallucinations remain the primary bottleneck~\cite{bai2024hallucination,wu2024faithfulness,zhong2024investigating,liu2025faithfulness,he-etal-2025-cracking}.

    \paragraph{Mainstream Strategies for Multimodal Reasoning.}
    To enhance multimodal reasoning, existing methods primarily focus on an RL paradigm that enables the autonomous refinement of reasoning trajectories via rollout sampling~\cite{shao2024deepseekmath,huang2025visionr1,shen2025vlmr1,li2025adacurl,he2026active}.
    Following this paradigm, a series of works have explored two kinds of improvements: including (1) training strategies like improving rollout diversity via mixing normal image and noisy augmentation~\cite{liu2025noisyrollout} or entropy-based regulation~\cite{cheng2025entropyreasoning} to balance exploration and exploitation; 
    (2) Visual-centric refinement like fine-grained visual enhancement via KL divergence comparison between normal and noisy images~\cite{huang2025spotlight, wang2025perception}, re-activating specific reasoning paths via introducing specific tokens~\cite{sun2025latentCoT,li2025lvr,yang2025mirage}, or verifying intermediate processes as the reward via auxiliary model~\cite{tian2025VAPO}.
    While effective, how to mitigate the need for external models and solely utilize the inherent states of modal abilities to achieve visual enhancement remains an open question.

    \paragraph{Visual Perception Methods.}

    Recent concurrent works have also explored enhancing visual perception in multimodal reasoning. 
    PEARL~\cite{zhang2025perceptual} and ViCrit~\cite{wang2025vicrit} enhance perception through external checklists or synthetic hallucination detection, but they require high computational or data construction costs. 
    On the other hand, SSL4RL~\cite{guo2025ssl4rl} and VisPlay~\cite{he2025visplay} utilize unlabeled data yet fail to maintain consistent visual attention during complex reasoning. 
    VGPO overcomes these trade-offs by deriving a Visual Focus Score directly from the internal hidden state. By enforcing continuous visual guidance via an attention compensation mechanism, VGPO mitigates visual decay and sparse activation. 
    This ensures robust reasoning and cross-domain generalizability, eliminating the reliance on external models or outcome-based proxy tasks found in prior work.
\section{Conclusion} 
    In this paper, we explore the critical limitation of insufficient visual activation and its temporal decay during multimodal reasoning. 
    To this end, we propose VGPO, a novel framework designed to alleviate this issue by leveraging intrinsic hidden states to autonomously ground visual focus. 
    By strategically coupling a Visual Attention Compensation mechanism with a dual-grained advantage re-weighting strategy, our method ensures consistent and sustained visual reasoning capability. 
    Comprehensive experiments validate that VGPO significantly enhances visual activation and delivers state-of-the-art performance.

\section*{Limitations}
    While VGPO effectively elevates visual expectations to amplify activation and mitigate forgetting during reasoning, we identify several avenues for future improvement:
    \begin{itemize}[leftmargin=*,itemsep=0.5pt, topsep=1.5pt]
        \item The strategy of progressively elevating visual expectations acts as a robust heuristic that may not represent the globally optimal solution for all reasoning patterns. For instance, in scenarios where the final reasoning steps rely strictly on logical deduction or calculation independent of visual cues, a mandated high visual focus might be less critical. Nevertheless, we regard this work as a pivotal milestone in identifying and addressing the overlooked issue of visual decay. We hope this study serves as a foundation for future research to explore more adaptive mechanisms that can dynamically rectify visual reliance according to the specific context of each reasoning step.
        \item As VGPO is designed to elicit and sustain vision intrinsic to the VLMs, its performance upper bound is inherently constrained by the representational quality of the visual encoder and projector. If the base model fails to encode critical visual features into the hidden states initially, our re-weighting strategy may not effectively recover this missing information. Consequently, our framework focuses on optimizing the utilization of visual cues rather than enhancing the raw perceptual capabilities of visual encoder itself.
    \end{itemize}





\bibliography{custom}

\appendix
\newpage
\label{sec:appendix}
\section{Experimental Settings}
\label{sec:appendix_experimental_settings}

\subsection{Training Datasets}
\label{sec:appendix_experimental_settings_training_dataset}
    We mainly conduct our training experiments on the ViRL39K~\cite{wang2025vlrethinker} dataset, a verifiable instruction-tuning benchmark designed for vision-language reasoning. This dataset comprises approximately 39,000 high-quality samples synthesized from proprietary collections and various upstream datasets, including Llava-OneVision~\cite{li2025llavaonevision}, R1-OneVision~\cite{yang2025r1onevision}, MM-Eureka~\cite{meng2025mmeureka}, MM-Math~\cite{sun-etal-2024-mm}, M3CoT~\cite{chen-etal-2024-m3cot}, DeepScaleR~\cite{deepscaler2025}, and MV-Math~\cite{wang2025mv-math}. ViRL39K is distinguished by a rigorous filtering pipeline that removes unverifiable queries, thereby ensuring extensive coverage of topics ranging from general chart interpretation to complex STEM problem-solving.

\label{sec:appendix_experimental_settings_evaluation_baselines}
\subsection{Baselines}

\begin{itemize}[left=0pt]
    \item \textbf{ThinkLite-VL}~\cite{wang2025thinklite-vl}. ThinkLite-VL bypasses knowledge distillation in favor of an MCTS-guided selection strategy that curates a compact, high-quality dataset based on sample difficulty. By applying reinforcement fine-tuning to these challenging instances, the method attains superior visual reasoning results while reducing data requirements by an order of magnitude.
    \item \textbf{VL-Rethinker}~\cite{wang2025vlrethinker}. VL-Rethinker enhances VLMs' slow-thinking capabilities through a distillation-free reinforcement learning framework that integrates Selective Sample Replay (SSR) to stabilize training and Forced Rethinking to incentivize self-reflection. By rehearsing high-value experiences and triggering explicit verification processes, the method effectively mitigates GRPO limitations and fosters the internalization of deliberate reasoning patterns.
    \item \textbf{MM-Eureka}~\cite{meng2025mmeureka}. MMEureka advances multimodal reasoning by combining the MMK12 dataset with a Qwen2.5-VL-based GRPO pipeline that employs online filtering to ensure gradient efficacy. To address training instability in large models, it utilizes a two-stage strategy involving initial training on MMK12 without KL divergence followed by fine-tuning on Geo3k with KL regularization.
    \item \textbf{NoisyRollout}~\cite{liu2025noisyrollout}. NoisyRollout enhances visual reasoning and robustness by computing GRPO advantages from hybrid clean and distorted trajectories, restricting policy optimization to uncorrupted inputs. Furthermore, it incorporates a noise annealing schedule to gradually diminish distortion, thereby effectively balancing exploration and stability.
    \item \textbf{PAPO}~\cite{wang2025perception}. PAPO addresses perception errors by introducing an Implicit Perception Loss that maximizes the KL divergence between policy outputs on original and masked inputs to enforce visual grounding. To ensure training stability, it employs a Double Entropy Loss regularization, offering a model-agnostic framework that jointly optimizes perception and reasoning without requiring additional annotations or teacher models.
    \item \textbf{VPPO}~\cite{huang2025spotlight}. VPPO integrates visual perception into RL by quantifying token-level visual dependency via KL divergence between policy distributions on original and perturbed image inputs. Based on this metric, it employs advantage modulation and sparse gradient masking to concentrate optimization exclusively on pivotal visual tokens.
\end{itemize}

\begin{table}[t]
    \centering
    \resizebox{1.0\linewidth}{!}{
        \begin{tabular}{lc}
        \toprule
        \textbf{Parameter} & \textbf{Configuration} \\
        \midrule
        \multicolumn{2}{l}{{\textbf{General Settings}}} \\ 
        \midrule
        Rollout Number & 8 \\
        Learning Rate & 1e-6 \\
        Global Batch Size & 128 \\
        Rollout Batch Size & 512 \\
        Val Batch Size & 1024 \\
        Max Prompt Length & 4096 \\
        Max Response Length & 2048 \\
        Reward & Binary Accuracy \\ 
        GPU Usage & 8$\times$H20, 96G Memory \\
        \midrule
        \multicolumn{2}{l}{\textbf{Qwen2.5-VL-3B/7B on ViRL39K (39K samples)}} \\ 
        \midrule
        Training Episodes & 2 \\
        Total Optimization Steps & 150 \\
        \midrule
        \multicolumn{2}{l}{\textbf{Qwen2.5-VL-7B on Geo3K (2.1K samples)}} \\
        \midrule
        Training Episodes & 15 \\
        Total Optimization Steps & 60 \\ 
        \midrule
        \multicolumn{2}{l}{\textbf{Qwen2.5-VL-7B on MMK12 (6.4K samples)}} \\ 
        \midrule
        Training Episodes & 12 \\
        Total Optimization Steps & 120 \\ 
        \midrule
        \multicolumn{2}{l}{\textbf{Qwen2.5-VL-32B on ViRL39K (39K samples)}} \\ 
        \midrule
        Training Episodes & 2 \\
        Total Optimization Steps & 150 \\
        GPU Usage & 32$\times$H20, 96G Memory \\
        \bottomrule
        \end{tabular}
    }
    \caption{Experimental hyperparameter configurations of our VGPO across 3B, 7B, and 32B-based backbones and different scalable training datasets.}
    \label{tab:hyperparameters}
\end{table}

\label{sec:appendix_experimental_settings_evaluation_benchmarks}
\subsection{Evaluation Benchmarks}
\textbf{Mathematical \& Geometric Reasoning:}

\begin{itemize}[left=0pt]
    \item \textbf{MathVista~\cite{lu2024mathvista}.} MathVista is designed to integrate challenges across diverse mathematical and visual domains, necessitating both fine-grained visual perception and compositional reasoning. The benchmark consists of 6,141 examples amassed from a wide array of sources, including 28 existing multimodal mathematics datasets and 3 constructed datasets.
    \item \textbf{MathVerse~\cite{zhang2024MathVerse}.} MathVerse is designed to mitigate the textual bias in existing benchmarks by curating 2,612 high-quality problems that necessitate genuine visual interpretation. Each problem is manually reformulated into six distinct versions with varying multimodal information density, resulting in a comprehensive corpus of approximately 15,000 test samples.
    \item \textbf{We-Math~\cite{qiao-etal-2025-math}.} We-Math transcends conventional accuracy metrics to scrutinize the underlying problem-solving mechanisms of LMMs through a hierarchical framework of 6,500 visual tasks spanning 67 distinct knowledge concepts. By employing a decompositional methodology that reduces composite problems into atomic sub-tasks, the benchmark facilitates a granular four-dimensional evaluation distinguishing among Insufficient Knowledge, Inadequate Generalization, Complete Mastery, and Rote Memorization.
    \item \textbf{MMK12~\cite{meng2025mmeureka}.} MMK12 benchmark constitutes a comprehensive evaluation framework designed to assess multimodal reasoning capabilities at the K-12 educational level. Comprising 2,000 high-quality instances, the benchmark is stratified across four scientific disciplines, specifically Mathematics, Physics, Chemistry, and Biology, with 500 multimodal multiple-choice questions dedicated to each field.
    \item \textbf{GeoMath~\cite{zhougeomath}.} Designed to address scarcity of reasoning data in Earth observation, GeoMath comprises 3,773 high-quality queries derived from aerial imagery, encompassing six distinct mathematical subjects across 20 sub-topics. The visual data are acquired via proprietary drone flights capturing a wide range of altitudes and viewing angles.
    \item \textbf{Geometry3K~\cite{lu-etal-2021-inter}.} Geometry3K comprises 3,002 geometry problems enriched with dense annotations in formal language, serving as a challenging task for abstract problem understanding and symbolic reasoning, requiring axiomatic knowledge.
\end{itemize}

\begin{table*}[t]
    \setlength{\tabcolsep}{1.0mm}
    \renewcommand\arraystretch{1.1}
    \centering
    \resizebox{1.0\linewidth}{!}{
        \begin{tabular}{lcccccccccccc}
        \toprule
        {\multirow{2}{*}{Models}} & \multicolumn{6}{c}{General Mathematical \& Geometric Reasoning} & {\multirow{2}{*}{Avg-$_\mathrm{Math}$}} & \multicolumn{4}{c}{Vision-dependent Multimodal Reasoning} & {\multirow{2}{*}{Avg-$_\mathrm{Vision}$}} \\ 
        \cmidrule{2-7} 
        \cmidrule{9-12}
         & {MathVista} & {MathVerse} & {WeMath} & {MMK12} & {GeoMath} & {Geo3k} &  & {LogicVista} & {Counting} & {MMMU-Pro} & {MathVerse${_V}$} &  \\ 
        \midrule
        {DAPO Baseline} & 68.7 & 69.6 & 70.8 & 77.0 & 51.3 & 45.6 & \cellcolor[HTML]{EFEFEF}{63.8} & 47.4 & 85.5 & 39.0 & 66.6 & \cellcolor[HTML]{EFEFEF}{59.6} \\
        {\ \ + Intra-trajectory} & \textbf{74.1} & 71.1 & 69.8 & \textbf{81.7} & 54.3 & 45.6 & \cellcolor[HTML]{EFEFEF}{66.1} & 48.1 & 95.0 & 40.2 & 66.5 & \cellcolor[HTML]{EFEFEF}{62.5} \\  
        {\ \ + Inter-trajectory} & 72.1 & 68.8 & 69.9 & 80.7 & \textbf{54.9} & 45.3 & \cellcolor[HTML]{EFEFEF}{65.3} & 47.9 & \textbf{95.5} & 40.2 & 64.6 & \cellcolor[HTML]{EFEFEF}{62.0} \\
        \rowcolor[HTML]{EFEFEF}
        {\ \ + Intra- \& Inter-trajectory} & \textbf{74.1} & \textbf{71.6} & \textbf{72.5} & 81.5 & {54.3} & \textbf{45.8} & \textbf{66.6} & \textbf{49.4} & \textbf{95.5} & \textbf{40.5} & \textbf{67.6} & \textbf{63.3} \\
        \bottomrule
        \end{tabular}
    }
    \vspace{-0.1cm}
    \caption{The detailed ablation study on the impact of Intra- and Inter-trajectory re-weighting strategies.}
    \label{fig:appendix_ablation_module}
\end{table*}

\begin{table*}[t]
    \setlength{\tabcolsep}{1.0mm}
    \renewcommand\arraystretch{1.1}
    \centering
    \resizebox{1.0\linewidth}{!}{
        \begin{tabular}{lcccccccccccc}
        \toprule
        {\multirow{2}{*}{Models}} & \multicolumn{6}{c}{General Mathematical \& Geometric Reasoning} & {\multirow{2}{*}{Avg-$_\mathrm{Math}$}} & \multicolumn{4}{c}{Vision-dependent Multimodal Reasoning} & {\multirow{2}{*}{Avg-$_\mathrm{Vision}$}} \\ 
        \cmidrule{2-7} 
        \cmidrule{9-12}
         & {MathVista} & {MathVerse} & {WeMath} & {MMK12} & {GeoMath} & {Geo3k} &  & {LogicVista} & {Counting} & {MMMU-Pro} & {MathVerse${_V}$} &  \\ 
        \midrule
        {Qwen2.5-VL-7B} & {68.5} & {40.2} & {47.8} & {49.4} & {51.2} & {42.9} & \cellcolor[HTML]{EFEFEF}{50.0} & {45.2} & {76.5} & {36.4} & {36.6} & \cellcolor[HTML]{EFEFEF}{48.7} \\
        {\ \ + DAPO} & 68.7 & 69.6 & 70.8 & 77.0 & 51.3 & 45.6 & \cellcolor[HTML]{EFEFEF}{63.8} & 47.4 & 85.5 & 39.0 & 66.6 & \cellcolor[HTML]{EFEFEF}{59.6} \\
        {\ \ + DAPO w/ Entropy} & 70.2 & \textbf{72.1} & 70.8 & 80.9 & 53.9 & 45.6 & \cellcolor[HTML]{EFEFEF}{65.6} & 48.3 & 90.5 & 40.1 & \textbf{68.7} & \cellcolor[HTML]{EFEFEF}{61.9} \\
        {\ \ + DAPO w/ KL$_{perception}$} & 70.5 & 71.1 & 70.6 & \textbf{81.8} & 53.2 & \textbf{46.9} & \cellcolor[HTML]{EFEFEF}{65.7} & 48.8 & 88.5 & 40.2 & 67.6 & \cellcolor[HTML]{EFEFEF}{61.3} \\
        \rowcolor[HTML]{EFEFEF}
        {\ \ + VGPO (Ours)} & \textbf{74.1} & {71.6} & \textbf{72.5} & {81.5} & \textbf{54.3} & 45.8 & \cellcolor[HTML]{EFEFEF}{\textbf{66.6}} & \textbf{49.4} & \textbf{95.5} & \textbf{40.5} & {67.6} & \cellcolor[HTML]{EFEFEF}{\textbf{63.3}} \\
        \bottomrule
        \end{tabular}
    }
    \vspace{-0.1cm}
    \caption{The detailed comparisons with other advantage shaping strategies, including the Entropy-based~\cite{cheng2025entropyreasoning} and KL-based~\cite{huang2025spotlight} methods.}
    \label{fig:appendix_ablation_adv_shaping}
\end{table*}

\noindent\textbf{Vision-Dependent Multimodal Reasoning:}

\begin{itemize}[left=0pt]
    \item \textbf{LogicVista~\cite{xiao2024logicvista}.} LogicVista is an evaluation benchmark designed to assess the integrated logical reasoning abilities of VLMs in visual contexts. Addressing the limitations of prior work in systematically evaluating logical proficiency. The dataset consists of 448 multiple-choice questions, each densely annotated with the correct answer and a human-written explanation. 
    \item \textbf{Super-CLEVR~\cite{li2023super}.} Super-CLEVR constitutes a virtual benchmark designed to rigorously assess the out-of-distribution robustness and domain generalization of VQA models. Through controllable data synthesis, it decouples inherent multi-modal factors to facilitate the isolated analysis of four specific domain shifts: visual complexity, question redundancy, concept distribution, and concept compositionality.
    \item \textbf{MMMU-Pro~\cite{yue-etal-2025-mmmu}.} MMMU-Pro advances the MMMU benchmark by employing a stringent curation process that eliminates text-solvable queries and expands candidate options to rigorously evaluate intrinsic multimodal reasoning. Additionally, it introduces a vision-only paradigm where inquiries are embedded within images, thereby necessitating integrated visual and textual interpretation.
\end{itemize}

\section{More Experimental Results}

\subsection{Detailed Results of Re-weighting Strategy}
    
    Table~\ref{fig:appendix_ablation_module} provides the whole performance breakdown across all datasets, complementing the aggregated results in the main text.
    
    \noindent\textbf{Effectiveness of Intra-trajectory Strategy.}
    This mechanism significantly enhances general mathematical reasoning. It yields substantial gains on MathVista (+5.4\%) and MMK12 (+4.7\%), demonstrating that reinforcing self-consistency within individual reasoning paths is critical to re-focus visual activation for multimodal reasoning
    
    \noindent\textbf{Impact of Inter-trajectory Strategy.}
    The Inter-trajectory module excels in geometry and visual counting tasks. It achieves a peak score of 54.9\% on GeoMath and matches the top performance on the Counting dataset. These results suggest that cross-path comparison effectively mitigates visual hallucinations in object-centric scenarios.
    
    \noindent\textbf{Synergy of the Two Strategies.}
    Combining both modules yields the most robust performance across mathematical and vision-dependent categories. Notably on WeMath, the integrated approach reverses a slight decline caused by Intra-module alone, achieving a superior score of 72.5\%. Furthermore, the combined strategy secures top results on challenging benchmarks like LogicVista and MMMU-Pro. This confirms a synergistic effect, where the Intra-module refines logical consistency while the Inter-module broadens the reasoning scope.
    
\subsection{Detailed Results of Advantage Shaping Methods}
    
    Table~\ref{fig:appendix_ablation_adv_shaping} presents a fine-grained comparison of VGPO against established advantage shaping strategies. Consistent with the main results, VGPO exhibits superior performance across both mathematical and vision-centric benchmarks.

    \noindent\textbf{Robustness in General Reasoning.}
    In contrast to generic regularization methods like Entropy and KL, which rely on implicit constraints on policy divergence, VGPO explicitly leverages visual feedback. This mechanism translates into substantial gains in General Mathematical \& Geometric Reasoning. VGPO secures the highest scores on 4 out of 6 datasets, establishing significant leads on MathVista and MathVerse. This demonstrates that visually-guided optimization not only enhances reasoning stability but does so while preserving the general problem-solving proficiency.
    
    \noindent\textbf{Efficacy in Visual Perception.}
    Our detailed analysis highlights the pivotal role of VGPO in Vision-dependent Multimodal Reasoning. The performance gap is particularly evident in the \textit{Counting} dataset, a task requiring rigorous visual grounding. VGPO achieves 95.5\% accuracy, surpassing the Entropy and KL baselines by a significant margin. Similar dominance is observed on LogicVista. These results validate our hypothesis that explicitly reinforcing visual attention through VGPO is far more effective for complex multimodal tasks than relying on implicit regularization techniques.

\subsection{Analysis of Training Dynamics}
\begin{figure}[t]
    \centering
    \vspace{-0.2cm}
    \begin{subfigure}{0.48\linewidth}
        \includegraphics[width=\linewidth]{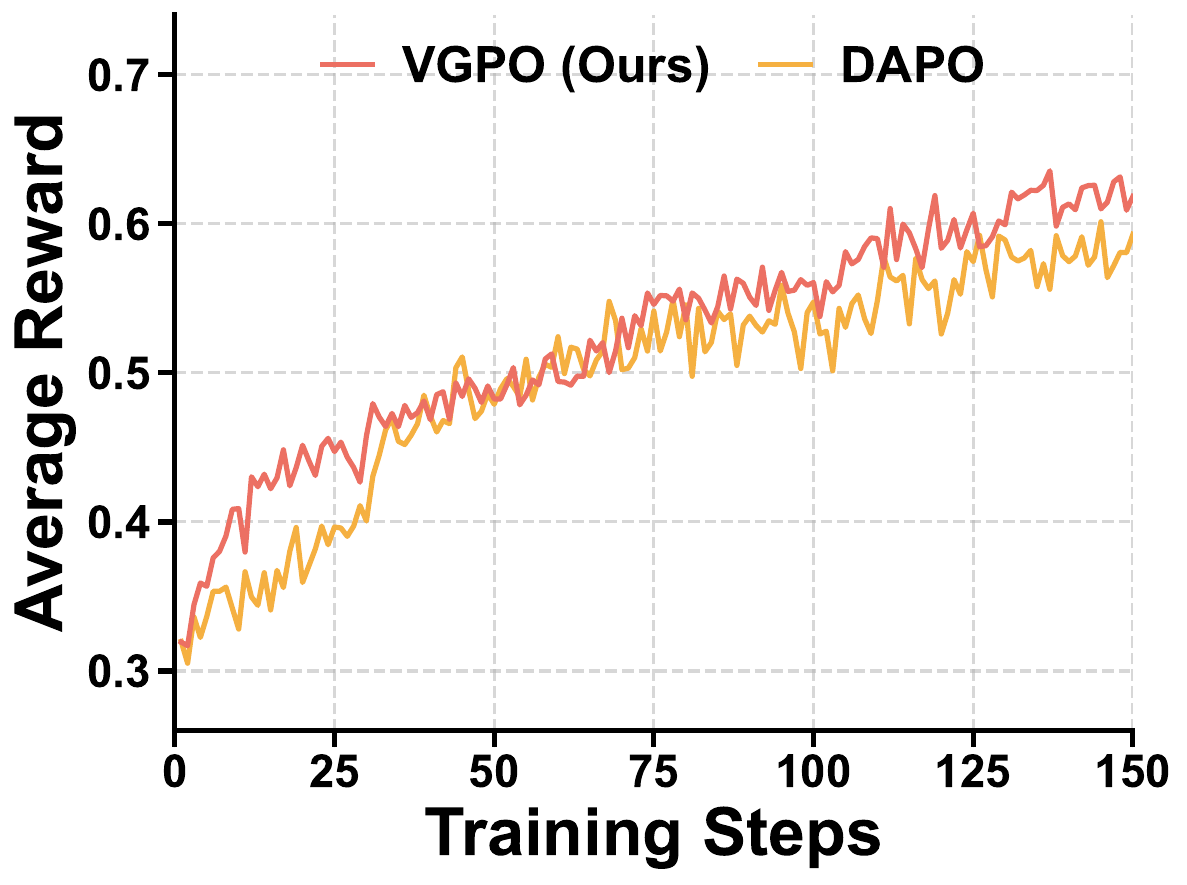}
        \subcaption{Training rewards.}
        \label{fig:training_dynamics_3b_a}
    \end{subfigure}
    \hspace{2pt}
    \begin{subfigure}{0.48\linewidth}
        \includegraphics[width=\linewidth]{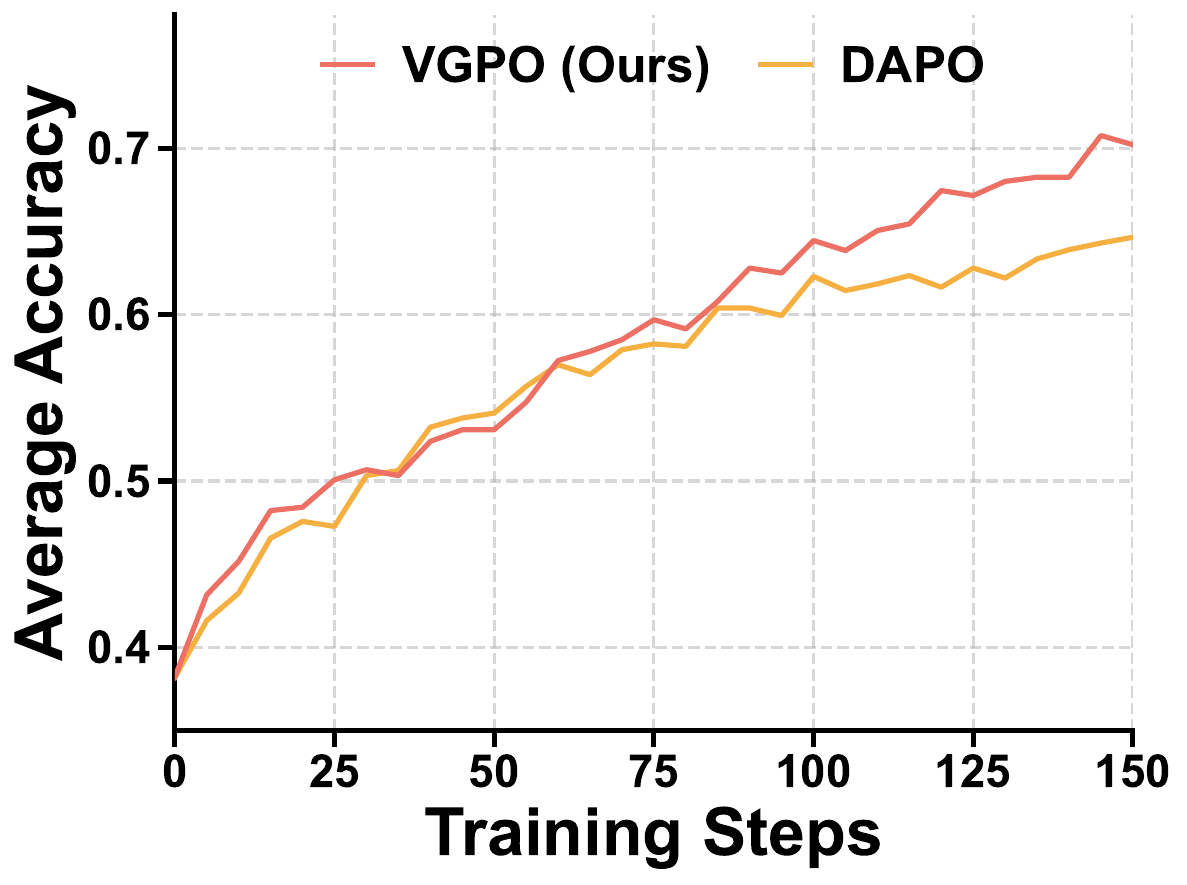}
        \subcaption{Validation accuracy.}
        \label{fig:training_dynamics_3b_b}
    \end{subfigure}
    \vspace{-0.6cm}
    \caption{Training dynamics of Qwen2.5-VL-3B: \textbf{(a)} training rewards and \textbf{(b)} validation accuracy on MMK12~\cite{meng2025mmeureka} across DAPO~\cite{yu2025dapo}, and our VGPO.}
    \label{fig:training_dynamics_3b}
\end{figure}

\begin{figure}[t]
    \centering
    \vspace{-0.2cm}
    \begin{subfigure}{0.48\linewidth}
        \includegraphics[width=\linewidth]{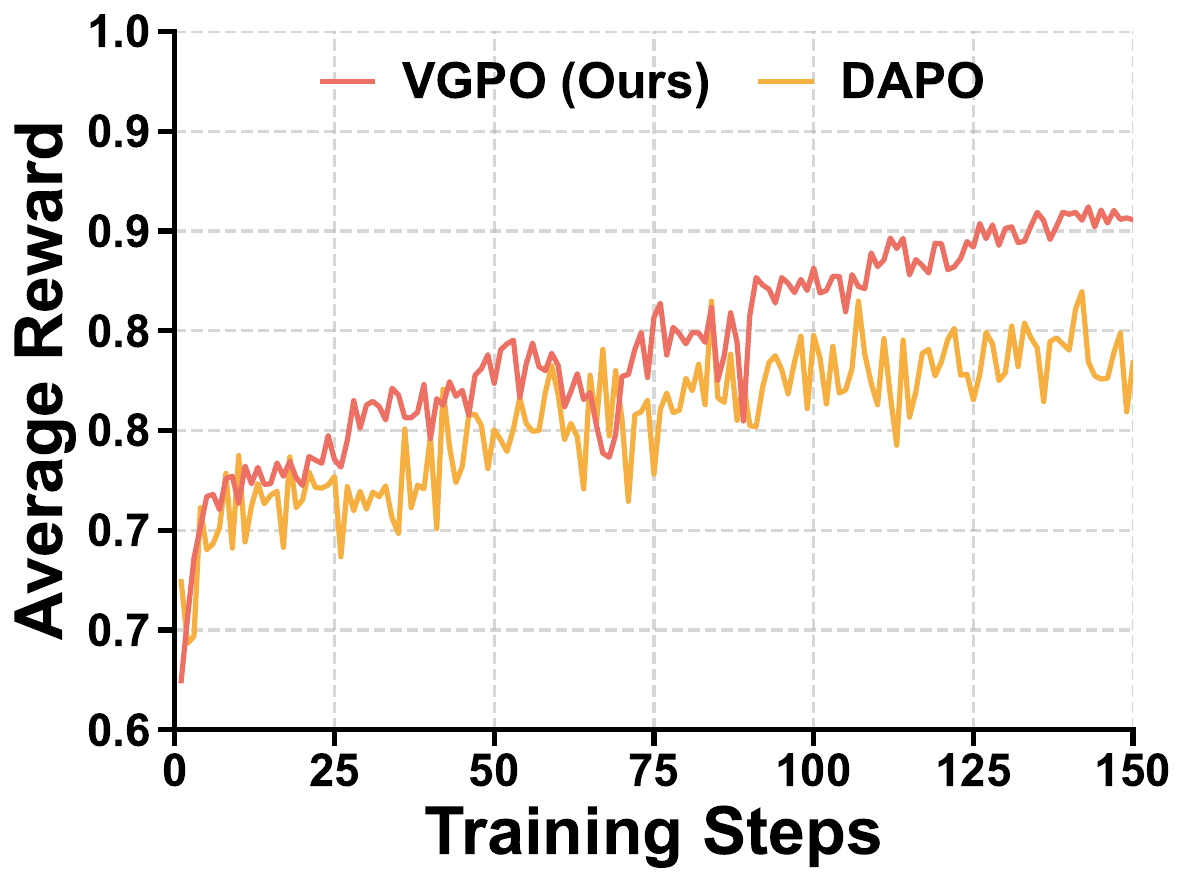}
        \subcaption{Training rewards.}
        \label{fig:training_dynamics_32b_a}
    \end{subfigure}
    \hspace{2pt}
    \begin{subfigure}{0.48\linewidth}
        \includegraphics[width=\linewidth]{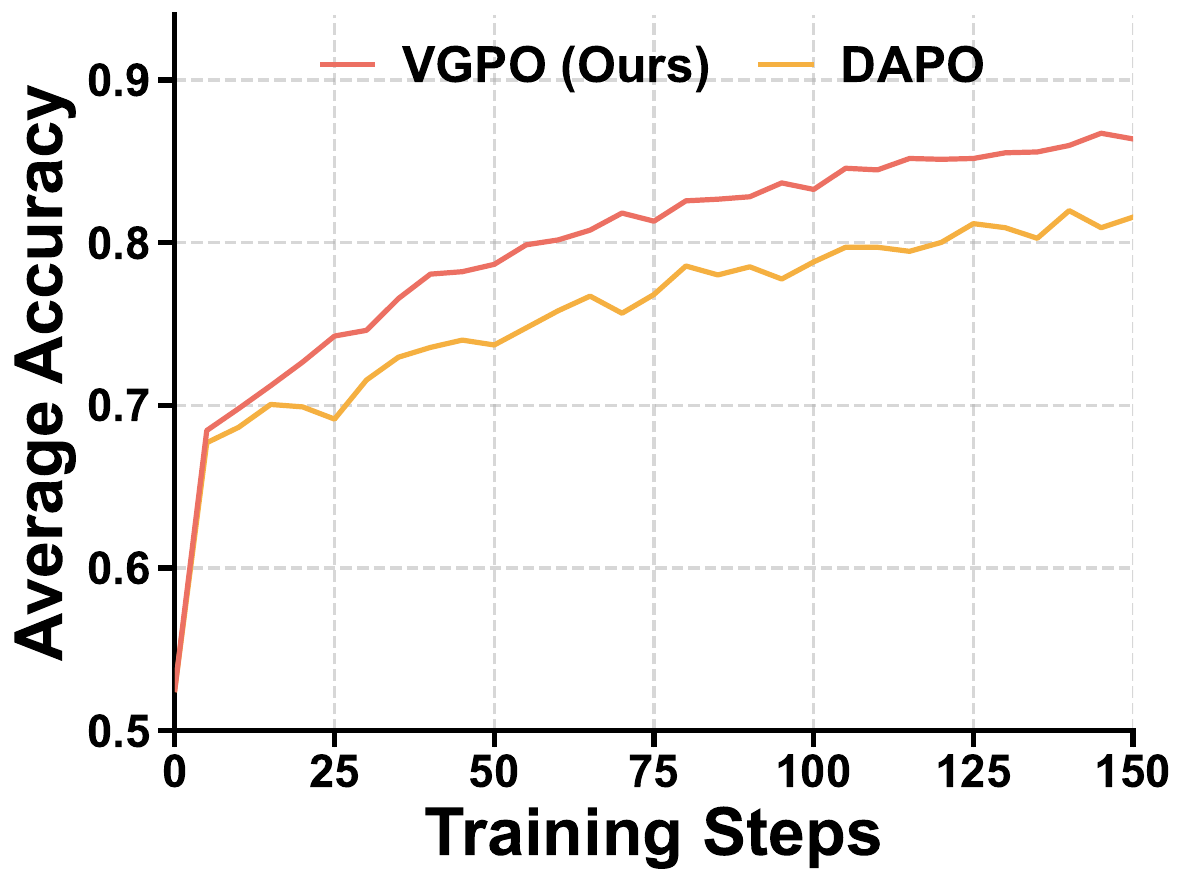}
        \subcaption{Validation accuracy.}
        \label{fig:training_dynamics_32b_b}
    \end{subfigure}
    \vspace{-0.6cm}
    \caption{Training dynamics of Qwen2.5-VL-32B: \textbf{(a)} training rewards and \textbf{(b)} validation accuracy on MMK12~\cite{meng2025mmeureka} across DAPO~\cite{yu2025dapo}, and our VGPO.}
    \label{fig:training_dynamics_32b}
\end{figure}

    To further investigate the convergence, stability, and scalability of our VGPO, we visualize the training dynamics of training rewards and validation accuracy on MMK12 dataset for both Qwen2.5-VL-3B (Figure~\ref{fig:training_dynamics_3b}) and Qwen2.5-VL-32B (Figure~\ref{fig:training_dynamics_32b}).

\noindent\textbf{Dynamics on Qwen2.5-VL-3B.}
    As shown in Figure~\ref{fig:training_dynamics_3b}, VGPO demonstrates a clear performance advantage. In terms of training rewards, VGPO maintains a consistently higher reward trajectory compared to DAPO throughout the training steps. More importantly, regarding validation accuracy, VGPO achieves a higher final accuracy, which indicates superior generalization capabilities.

\noindent\textbf{Dynamics on Qwen2.5-VL-32B.}
    The advantages of VGPO are even more pronounced on the larger 32B model, particularly regarding stability. Figure~\ref{fig:training_dynamics_32b}(a) highlights that DAPO suffers from severe oscillation during training, whereas VGPO exhibits a smooth and steady increase in rewards. This suggests that our visual-guided strategy effectively reduces variance and stabilizes the optimization process for large-scale models. Consistent with the rewards, Figure~\ref{fig:training_dynamics_32b}(b) shows that VGPO consistently outperforms DAPO in validation accuracy, establishing a significant margin by the end of training.

\subsection{Detailed Results of Different Compensation Schedules}
\label{sec:appendix_compensation_schedules}

    Table~\ref{tab:appendix_compensation_schedules} provides the detailed performance breakdown across all datasets, complementing the aggregated results in the main text.

    \begin{table*}[t]
    \setlength{\tabcolsep}{1.0mm}
    \renewcommand\arraystretch{1.1}
    \centering
    \resizebox{1.0\linewidth}{!}{
        \begin{tabular}{lcccccccccccc}
        \toprule
        {\multirow{2}{*}{Compensation Schedules}} & \multicolumn{6}{c}{General Mathematical \& Geometric Reasoning} & {\multirow{2}{*}{Avg-$_\mathrm{Math}$}} & \multicolumn{4}{c}{Vision-dependent Multimodal Reasoning} & {\multirow{2}{*}{Avg-$_\mathrm{Vision}$}} \\ 
        \cmidrule{2-7} 
        \cmidrule{9-12}
         & {MathVista} & {MathVerse} & {WeMath} & {MMK12} & {GeoMath} & {Geo3k} &  & {LogicVista} & {Counting} & {MMMU-Pro} & {MathVerse${_V}$} &  \\ 
        \midrule
        {Baseline (DAPO)} & 68.7 & 69.6 & 70.8 & 77.0 & 51.3 & 45.6 & \cellcolor[HTML]{EFEFEF}{63.8} & 47.4 & 85.5 & 39.0 & 66.6 & \cellcolor[HTML]{EFEFEF}{59.6} \\
        {w/ Step-Function} & 69.1 & 69.2 & 69.4 & 80.3 & 52.3 & \textbf{47.6} & \cellcolor[HTML]{EFEFEF}{64.7} & 47.5 & 91.5 & 39.2 & 64.7 & \cellcolor[HTML]{EFEFEF}{60.7} \\
        {w/ Exponential} & 70.6 & 69.9 & 69.3 & 81.4 & 54.2 & 44.9 & \cellcolor[HTML]{EFEFEF}{65.1} & 49.0 & 90.5 & 38.8 & 65.5 & \cellcolor[HTML]{EFEFEF}{61.0} \\
        \rowcolor[HTML]{EFEFEF}
        {\textbf{w/ Linear (Ours)}} & \textbf{74.1} & \textbf{71.6} & \textbf{72.5} & \textbf{81.5} & \textbf{54.3} & 45.8 & \textbf{66.6} & \textbf{49.4} & \textbf{95.5} & \textbf{40.5} & \textbf{67.6} & \textbf{63.3} \\
        \bottomrule
        \end{tabular}
    }
    \vspace{-0.1cm}
    \caption{Results of General Mathematical \& Geometric Reasoning and Vision-dependent Multimodal Reasoning tasks with different compensation schedules. \emph{Note}: For the Exponential schedule, we test powers of 1.0 and 2.0 (reporting the best result from 2.0). We test 1.0 for the Step-wise schedule.}
    \label{tab:appendix_compensation_schedules}
\end{table*}

    \begin{table*}[t]
    \setlength{\tabcolsep}{1.0mm}
    \renewcommand\arraystretch{1.1}
    \centering
    \resizebox{1.0\linewidth}{!}{
        \begin{tabular}{lcccccccccccc}
        \toprule
        {\multirow{2}{*}{Models}} & \multicolumn{6}{c}{General Mathematical \& Geometric Reasoning} & {\multirow{2}{*}{Avg-$_\mathrm{Math}$}} & \multicolumn{4}{c}{Vision-dependent Multimodal Reasoning} & {\multirow{2}{*}{Avg-$_\mathrm{Vision}$}} \\ 
        \cmidrule{2-7} 
        \cmidrule{9-12}
         & {MathVista} & {MathVerse} & {WeMath} & {MMK12} & {GeoMath} & {Geo3k} &  & {LogicVista} & {Counting} & {MMMU-Pro} & {MathVerse${_V}$} &  \\ 
        \midrule
        {Baseline (DAPO, Qwen2.5-VL-7B)} & 68.7 & 69.6 & 70.8 & 77.0 & 51.3 & 45.6 & \cellcolor[HTML]{EFEFEF}{63.8} & 47.4 & 85.5 & 39.0 & 66.6 & \cellcolor[HTML]{EFEFEF}{59.6} \\
        \rowcolor[HTML]{EFEFEF} {\textbf{+VGPO (Ours)}} & \textbf{74.1} & \textbf{71.6} & \textbf{72.5} & \textbf{81.5} & \textbf{54.3} & \textbf{45.8} & \cellcolor[HTML]{EFEFEF}{\textbf{66.6\textcolor{tablecolor}{$^{\triangle  4.4\%}$}}} & \textbf{49.4} & \textbf{95.5} & \textbf{40.5} & \textbf{67.6} & \cellcolor[HTML]{EFEFEF}{\textbf{63.3\textcolor{tablecolor}{$^{\triangle  6.2\%}$}}} \\
        \midrule
        {Baseline (DAPO, Qwen2.5-VL-3B)} & 63.9 & 57.1 & \textbf{62.9} & 64.1 & 47.6 & \textbf{36.3} & \cellcolor[HTML]{EFEFEF}{55.3} & 43.0 & 65.5 & 31.5 & 53.1 & \cellcolor[HTML]{EFEFEF}{48.3} \\
        \rowcolor[HTML]{EFEFEF} {\textbf{+VGPO (Ours)}} & \textbf{65.0} & \textbf{61.4} & 62.1 & \textbf{71.6} & \textbf{50.2} & 36.1 & \cellcolor[HTML]{EFEFEF}{\textbf{57.7\textcolor{tablecolor}{$^{\triangle  4.3\%}$}}} & \textbf{45.9} & \textbf{78.0} & \textbf{32.5} & \textbf{58.1} & \cellcolor[HTML]{EFEFEF}{\textbf{53.6\textcolor{tablecolor}{$^{\triangle  11.0\%}$}}} \\
        \midrule
        {Baseline (DAPO, Qwen2-VL-2B)} & 50.8 & 33.2 & 36.3 & 42.3 & 34.8 & 8.7 & \cellcolor[HTML]{EFEFEF}{34.4} & 30.0 & 66.5 & 19.8 & 31.6 & \cellcolor[HTML]{EFEFEF}{37.0} \\
        \rowcolor[HTML]{EFEFEF} {\textbf{+VGPO (Ours)}} & \textbf{51.1} & \textbf{37.8} & \textbf{42.5} & \textbf{44.2} & \textbf{37.0} & \textbf{16.8} & \cellcolor[HTML]{EFEFEF}{\textbf{38.2\textcolor{tablecolor}{$^{\triangle  11.1\%}$}}} & \textbf{30.2} & \textbf{83.5} & \textbf{21.9} & \textbf{36.4} & \cellcolor[HTML]{EFEFEF}{\textbf{43.0\textcolor{tablecolor}{$^{\triangle  16.2\%}$}}} \\
        \bottomrule
        \end{tabular}
    }
    \vspace{-0.1cm}
    \caption{Performance comparison of stronger to weaker visual encoders on General Mathematical \& Geometric Reasoning and Vision-dependent Multimodal Reasoning tasks.}
    \label{tab:appendix_weaker_encoder}
\end{table*}

    We adopt the Linear Compensation Strategy since it aligns well with the progressive, continuous, and nearly linear decay of visual attention observed in our empirical analysis across four visual-dependent tasks (LogicVista, CLEVR Counting, MMMU-Pro, MathVerse-V).
    To explore potential alternatives, we compare our Linear strategy against Step-Function and Exponential schedules. As shown in Table~\ref{tab:appendix_compensation_schedules}, all these schedules exhibit higher performance compared to the baseline. However, the Linear strategy consistently achieves the best performance. We attribute these results to the following reasons:

    \begin{itemize}[left=0pt]
        \item \textbf{Exponential schedule (over-correction):} This schedule puts most of the bonus on the very last tokens, which often correspond to calculation or formatting rather than visual grounding. Over-weighting those tokens dilutes the benefit of compensating truly visual-dependent steps and can hurt accuracy compared to the more balanced linear schedule.
        \item \textbf{Step-Function schedule (training instability):} This schedule makes position compensation jump from zero to full compensation at a single threshold. However, it does not match our observed progressive forgetting: visual attention decays gradually over the trajectory, so we use a linear schedule that increases compensation smoothly rather than a single step.
    \end{itemize}

\subsection{Robustness on Weaker Visual Encoders}
\label{sec:appendix_weaker_encoders}

    To investigate the robustness of VGPO across different visual encoder capacities, we evaluate our method using Qwen2-VL-2B-Instruct~\cite{bai2025qwen25vl}, which employs a significantly smaller and weaker visual encoder compared to the Qwen2.5-VL series. As shown in Table~\ref{tab:appendix_weaker_encoder}, despite the inherently constrained perceptual capacity of the 2B baseline, applying VGPO achieves even better relative improvements (+11.1\% on Math and +16.2\% on vision-dependent tasks). This empirical evidence confirms that VGPO demonstrates strong generalizability and robustness, effectively resolving the policy-level bottleneck of temporal visual forgetting regardless of the base visual encoder's strength.

\subsection{Detailed Results of Full-trajectory Compensation}
\label{sec:appendix_full_trajectory}

    Table~\ref{tab:appendix_full_trajectory} provides the detailed performance breakdown across all datasets, complementing the aggregated results in the main text. As shown in Table~\ref{tab:appendix_full_trajectory}, we find that the full-trajectory compensation strategy significantly drops performance on the majority of benchmarks. Moreover, the Late/Early Visual Activation Ratio exhibits a lower value (0.8591) than our VGPO (1.0696). This indicates that excessive elevation of early-trajectory compensation often overemphasizes early visual attention and leads to higher visual forgetting with lower performance.

    \begin{table*}[t]
    \setlength{\tabcolsep}{1.0mm}
    \renewcommand\arraystretch{1.1}
    \centering
    \resizebox{1.0\linewidth}{!}{
        \begin{tabular}{lcccccccccccc}
        \toprule
        {\multirow{2}{*}{\begin{tabular}{c}Compensation\\Schedules \end{tabular}}} & \multicolumn{6}{c}{General Mathematical \& Geometric Reasoning} & {\multirow{2}{*}{Avg-$_\mathrm{Math}$}} & \multicolumn{4}{c}{Vision-dependent Multimodal Reasoning} & {\multirow{2}{*}{Avg-$_\mathrm{Vision}$}} \\ 
        \cmidrule{2-7} 
        \cmidrule{9-12}
         & {MathVista} & {MathVerse} & {WeMath} & {MMK12} & {GeoMath} & {Geo3k} &  & {LogicVista} & {Counting} & {MMMU-Pro} & {MathVerse${_V}$} &  \\ 
        \midrule
        {Baseline (DAPO)} & 68.7 & 69.6 & 70.8 & 77.0 & 51.3 & 45.6 & \cellcolor[HTML]{EFEFEF}{63.8} & 47.4 & 85.5 & 39.0 & 66.6 & \cellcolor[HTML]{EFEFEF}{59.6} \\
        {Full-trajectory} & 71.4 & 50.3 & 53.8 & 60.1 & 44.9 & 37.3 & \cellcolor[HTML]{EFEFEF}{53.0} & 40.5 & 92.0 & 36.4 & 47.8 & \cellcolor[HTML]{EFEFEF}{54.2} \\
        \rowcolor[HTML]{EFEFEF}
        {\textbf{Late-trajectory (Ours)}} & \textbf{74.1} & \textbf{71.6} & \textbf{72.5} & \textbf{81.5} & \textbf{54.3} & \textbf{45.8} & \textbf{66.6} & \textbf{49.4} & \textbf{95.5} & \textbf{40.5} & \textbf{67.6} & \textbf{63.3} \\
        \bottomrule
        \end{tabular}
    }
    \vspace{-0.2cm}
    \caption{Performance comparison of Full-trajectory compensation versus our Late-trajectory compensation strategy.}
    \label{tab:appendix_full_trajectory}
\end{table*}

\subsection{Prompt Template} \label{sec:prompt_template}
    \begin{tcolorbox}[colback=black!5!white,colframe=cyan!10!black,title=Template for Multi-modal Reasoning]
    \textbf{SYSTEM} \\
    You are a helpful assistant.\\
    
    \textbf{USER} \\
    <question> \\
    You first think through the reasoning process as an internal monologue, enclosed within <think> </think> tags. Then, provide your final answer enclosed within \textbackslash boxed\{\}.
    \end{tcolorbox}

\section{Discussion about Visual Focus Score \emph{vs.} Attention Weights}
\label{sec:appendix_discussion}

    To validate the theoretical grounding of our proposed Visual Focus Score, we conduct an empirical correlation analysis between the Visual Focus Score and actual attention weights. 

    \noindent\textbf{Empirical validation of correlation with Attention Weights.}
    We select the validation dataset (MMK12-val~\cite{meng2025mmeureka}) for this analysis. At each generation step of each sample:
    \begin{itemize}[left=0pt]
        \item For the \textbf{Visual Focus Score}, we compute the cosine similarity between the final hidden state of the $t$-th generated token and the mean-pooled visual prototype.
        \item For the \textbf{Attention Weights}, we compute the sum of the attention probabilities from the $t$-th generated token to all image token positions, averaged across all attention heads in the final layer.
    \end{itemize}
    We then calculate the Pearson correlation coefficient between these two step-wise sequences ($t=1,2,\dots,T$). We observe a strong positive correlation ($r \approx 0.67$), confirming that our similarity-based Visual Focus Score closely aligns with the actual visual attention mechanism.

    \noindent\textbf{Why use hidden-state similarity rather than actual attention weights?}
    During implementation, extracting actual attention weights requires setting \texttt{output\_attentions=True}. This causes the model to fall back to the less efficient eager attention implementation rather than Flash Attention~\cite{shah2024flashattention}, losing Flash Attention's speed and memory benefits (typically incurring a 20\%--30\% additional computational cost). In contrast, our hidden-state-based implementation only requires \texttt{output\_hidden\_states=True}, which adds approximately 5\%--10\% overhead and remains fully compatible with Flash Attention, making it highly efficient for large-scale training.

\section{The Use of LLMs}
    This manuscript utilizes LLMs exclusively for the purpose of linguistic refinement and stylistic improvement to enhance clarity. The LLMs serve strictly as a writing aid and contribute no intellectual content to the research conceptualization, methodological design, data execution, analysis, or interpretation. These core scientific components remain the sole work of the authors. All suggestions provided by the LLMs are critically reviewed and selectively adopted to ensure the accuracy and integrity of the text. The authors retain full accountability for the content and findings presented herein, and the LLM does not qualify for authorship or contributorship.

\section{Failure Case Analysis}
\label{sec:appendix_failure_case}

    To provide deeper insights into the boundaries of our proposed VGPO, we present both qualitative and quantitative analyses of failure cases where increased visual focus does not translate to improved accuracy.

    \noindent\textbf{Qualitative insight (hallucination):}
    Our error analysis reveals that if the base visual encoder fundamentally misinterprets a feature (\eg, misrepresenting a blurry ``6'' as an ``8'' due to low resolution), VGPO can actually reinforce this error. By incentivizing the LLM to strongly attend to a flawed visual embedding, the model produces confidently incorrect answers, rather than falling back on common-sense priors that might have occasionally guessed correctly.

    \noindent\textbf{Quantitative insight:}
    As shown in our original analysis in Figure~\ref{fig02:analysis}(c), while correct samples generally show higher visual accumulation, there is a subset of incorrect samples that also exhibit high visual focus scores. This quantitatively demonstrates that ``looking closely'' is a necessary but not sufficient condition for correctness. The model must also possess the correct world knowledge and reasoning logic to interpret the visual cues properly. Here we provide two typical failure cases to illustrate these points.

    \newpage
    \begin{tcolorbox}[colback=black!5!white,colframe=cyan!10!black,title={Failure Case 1: Visual Perception Error (GeoMath, id: geomath-2884)}]
    \begin{center}
        \includegraphics[width=0.95\linewidth]{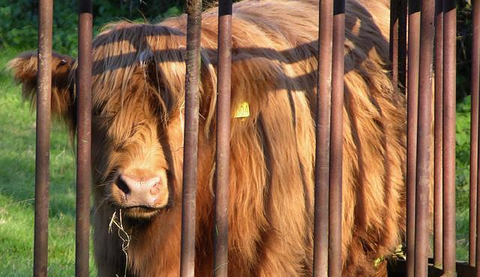}
    \end{center}
    \textbf{Question}: \texttt{<image>} How many vertical bars are visible in the image, and what is the total number of elements (cow and bars) present in the image? \\
    \textbf{Ground Truth}: 8 \\
    \textbf{Model Prediction}: To determine the number of vertical bars and the total number of elements in the image, let's break it down step by step. \\
    1. Count the vertical bars: The image shows a series of vertical bars that form a fence. By visually inspecting the image, it appears there are at least 10 vertical bars. (\textit{visual counting error}) \\
    2. Count the elements: There is one cow visible in the image. There are 10 vertical bars. \\
    3. Calculate the total number of elements: Total elements = 10 + 1 = 11. \\
    Therefore, the total number of elements is: 
    \textbackslash boxed\{11\}. \\
    \textbf{Failure Analysis}: The model's prediction fails due to an Object Counting Error. It correctly identifies the task of counting bars and adding the cow, but falters on the initial visual perception step, incorrectly counting 10 vertical bars instead of the actual visible bars (7, not been occluded by other bars or the cow). While its mathematical logic ($10 + 1 = 11$) is internally consistent, the entire reasoning is built on a flawed premise. The root cause is a failure in basic visual perception, not a flaw in logical deduction.
    \end{tcolorbox}

    \begin{tcolorbox}[colback=black!5!white,colframe=cyan!10!black,title={Failure Case 2: Logical Reasoning Error (MathVerse-V, id: 2)}]
    \begin{center}
        \includegraphics[width=0.95\linewidth]{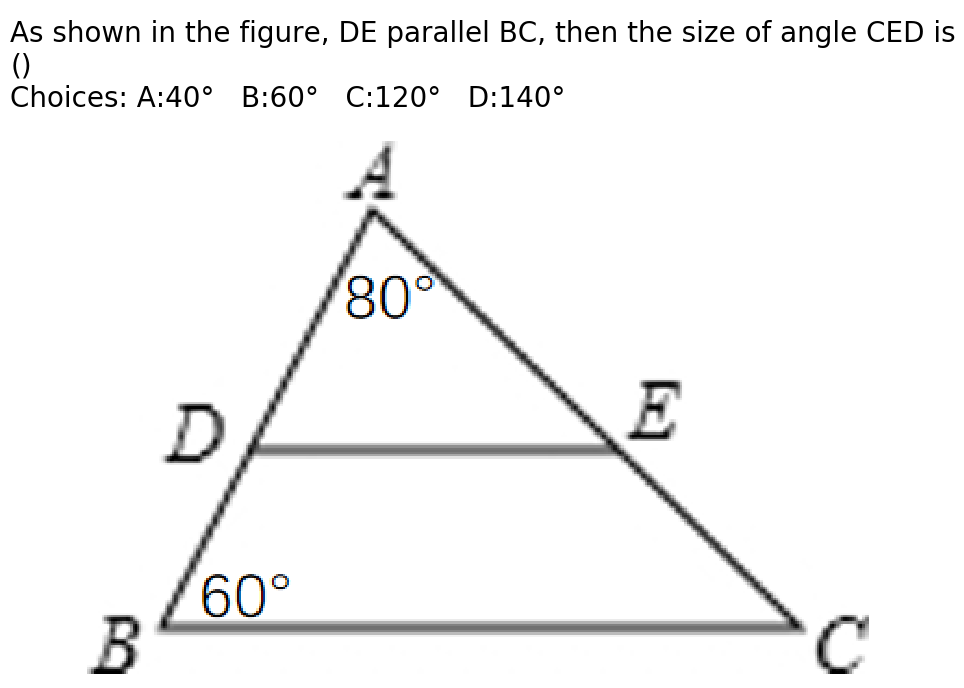}
    \end{center}
    \textbf{Question}: \texttt{<image>}. \\
    \textbf{Ground Truth}: D \\
    \textbf{Model Prediction}: To determine the size of angle $\text{CED}$, we start by analyzing the given information. First, we know that $\angle B = 60^\circ$ and $\angle A = 80^\circ$. Therefore, we can find $\angle C$ using the fact that the sum of the angles in a triangle is $180^\circ$:
    \begin{align*}
        \angle C &= 180^\circ - \angle A - \angle B \\
                 &= 180^\circ - 80^\circ - 60^\circ = 40^\circ
    \end{align*}
    Since $DE \parallel BC$, by the corresponding angles property (\textit{wrong geometric theorem}), we have:
    $$\angle CED = \angle C = 40^\circ$$
    Therefore, the size of angle $\text{CED}$ is $40^\circ$. Thus, the correct answer is: \textbackslash boxed\{A\}. \\
    \textbf{Failure Analysis}: The model correctly perceives the geometric setup and calculates $\angle C$ as $40^\circ$. The error stems from flawed knowledge: it misapplies the parallel line theorem, incorrectly equating $\angle CED$ with $\angle C$. It confuses supplementary consecutive interior angles with equal corresponding angles, showing that while the model ``looks'' correct, it lacks the proper logic to interpret what it sees.
    \end{tcolorbox}

\end{document}